# Advanced Assistance for Traffic Crash Analysis: An AI-Driven Multi-Agent Approach to Pre-Crash Reconstruction

Gerui Xu[a], Boyou Chen[a], Huizhong Guo[b], Dave LeBlanc[b], Arpan Kusari[b], Efe Yarbasi[b], Ananna Ahmed[c], Zhaonan Sun[c], Shan Bao[a, b*]

[a]*Industrial and Manufacturing Systems Engineering Department, University of Michigan-Dearborn, 4901 Evergreen Rd, Dearborn, MI, 48128, USA*

[b]*University of Michigan Transportation Research Institute, 2901 Baxter Road, Ann Arbor, MI, 48109, USA*

[c]*Toyota Motor Engineering & Manufacturing North America, Inc., USA*

**ABSTRACT**

Traffic collision reconstruction traditionally relies on human expertise and, when performed properly, can be incredibly accurate. However, attempting to perform pre-crash reconstruction, i.e., reconstructing the driver and vehicle behaviors that preceded the actual crash, poses significantly more challenges. This study develops a multi-agent AI framework that reconstructs pre-crash scenarios and infers vehicle behaviors from fragmented collision data. We present a two-phase collaborative framework combining reconstruction and reasoning phases. The system processes 277 rear-end lead vehicle deceleration (LVD) collisions from the Crash Investigation Sampling System (CISS, 2017-2022), integrating textual crash reports, structured tabular data, and visual scene diagrams. Phase I generates natural-language crash reconstructions from multimodal inputs. Phase II performs in-depth crash reasoning by combining these reconstructions with the temporal Event Data Recorder (EDR). This enables precise identification of striking and struck vehicles while isolating the EDR records most relevant to the collision moment, thereby revealing crucial pre-crash driving behaviors. For validation, we applied it to all LVD cases, focusing on a subset of 39 complex crashes where multiple EDR records per collision introduced ambiguity (e.g., due to missing or conflicting data). Ground truth was established via consensus between manual annotations (two independent researchers), with a separate large language model used only to flag possible conflicts for re-checking.

The framework achieved 100% accuracy across all 4,155 trials, with three reasoning models producing identical outputs, confirming that performance derives from the structured prompt design rather than model-specific characteristics. In contrast, research analysts without specialized reconstruction training achieved 92.31% accuracy on the same 39 complex cases. Ablation experiments further demonstrated that removing the structured reasoning anchors reduced case-level accuracy from 99.7% to 96.5%, with errors spreading from a single output type to multiple analytical dimensions. The system maintained robust performance even when processing incomplete data. This zero-shot evaluation, conducted without any domain-specific training or fine-tuning, demonstrates that the framework's effectiveness stems from its multi-agent architecture and prompt engineering, offering a scalable approach for AI-assisted pre-crash analysis.

## 1 INTRODUCTION

Road traffic crashes remain a critical public health and safety challenge worldwide. Every year, over a million people lose their lives in roadway accidents, with tens of millions more injured or disabled [1]. In the United States, the 2025 National Highway Traffic Safety Administration (NHTSA) report indicates that 40,901 people

* Corresponding author.

E-mail addresses: **geruixu@umich.edu** (Gerui Xu), **Boyou@umich.edu** (Boyou Chen), **shanbao@umich.edu** (Shan Bao)

died in motor vehicle crashes in 2023 [2]. Between 2020 and 2024, Michigan recorded 1,398,246 motor vehicle collisions, of which 867,364 (62.03%) involved two or more vehicles. Rear-end crashes totaled 322,270 cases, representing 37.16% of crashes involving two or more vehicles (Data source from Michigan Traffic Crash Facts (MTCF): **https://www.michigantrafficcrashfacts.org/**). Consequently, a comprehensive understanding of pre-crash dynamics in these incidents is critical for developing effective prevention strategies and advancing vehicle safety systems.

However, traditional pre-crash reconstruction methods face substantial limitations when analyzing complex collision scenarios. Current approaches depend heavily on human experts to manually integrate fragmented evidence from multiple sources—witness accounts, physical evidence, vehicle damage, and Event Data Recorder (EDR) information [3-6]. This process becomes particularly challenging with incomplete data, ambiguous EDR records, and diverse multimodal information including crash narratives, vehicle dynamics, and scene diagrams [7]. The cognitive burden of processing such heterogeneous information may lead to analytical inconsistencies and reduced accuracy in reconstructing critical pre-crash dynamics. These limitations highlight the opportunity for sophisticated analytical frameworks capable of processing fragmented multimodal data to help reconstruct the critical pre-crash time window, where key driver behaviors and vehicle dynamics ultimately determine collision outcomes.

### *1.1 Pre-crash Reconstruction Challenges*

Traditional traffic accident reconstruction relies on experts analyzing fragmented evidence to determine causal relationships and liability [8, 9]. The core framework of accident reconstruction remains based on expert experience and rule-based methods, integrating and analyzing limited investigative evidence (such as scene investigation, road environment, eyewitness testimony, and other data), with experts ultimately responsible for interpreting the significance of findings, weighing conflicting evidence, and making judgments regarding crash causation and liability [10, 11]. However, to truly understand the causal mechanisms of accidents and formulate proactive prevention strategies, researchers and industry practitioners require deeper insights into driver decision-making and behavioral patterns prior to collision events. Unlike post-crash reconstruction that relies on solidified physical evidence, pre-crash reconstruction needs to infer the dynamic evolution of vehicle trajectories from limited and indirect evidence. Analysts are required to synthesize multiple heterogeneous and fragmented sources of evidence (eyewitness statements, crash reports, onboard sensor data, visual evidence), and under conditions where information may be incomplete or even contradictory, construct a coherent and interpretable logical sequence of crash evolution from numerous possible hypotheses.

The evidence used in pre-crash reconstruction can generally be categorized into three types, each with inherent limitations. The first type is **human-recorded information**, including witness statements and official crash reports. Witness statements are subject to memory decay, limited vantage points, and psychological stress, which compromise the accuracy and completeness of event descriptions [12]. Official crash reports have been widely documented to suffer from misreporting, underreporting, and internal contradictions across different countries [13-15]. Studies have shown that alcohol-related factors are omitted in 30%–50% of fatal or serious injury crashes in certain U.S. states [15, 16] , and even the widely adopted CISS dataset has been found to contain systematic biases and sampling errors that may distort crash severity estimates [17]. To compensate for these deficiencies, analysts often resort to manual interpretation of narrative records or cross-validation across data sources, yet this process itself introduces additional uncertainty and subjectivity [18]. The second type is **scene evidence**, including skid marks, pavement gouges and scratches, vehicle damage distributions, and roadway environmental conditions. These physical traces may be altered or lost due to changing weather conditions, subsequent traffic interference, or delayed emergency response [3, 6]. The third type is **onboard time-series data**, which provides a more objective dimension for pre-crash dynamic analysis, but whose effective use also presents notable challenges. Taking the most widely deployed EDR as an example, its recording window typically covers only the final 5–10 seconds before a crash, which is often insufficient to fully trace the initial conditions that precipitated the event [19]. CAN bus data may offer a longer temporal window but requires specialized decoding, while dashcam video provides continuous visual information but lacks precise kinematic parameters.

The central challenge of pre-crash analysis is to achieve temporal alignment and semantic fusion of heterogeneous evidence under conditions of missing or contradictory data. Domain experts can accomplish this through experience and physical intuition, but manual interpretation struggles to maintain both efficiency and consistency across large case volumes. There is therefore a pressing need for intelligent systems with inferential capabilities that can more efficiently, objectively, and consistently evaluate all plausible evidence-chains, thereby assisting human experts in improving analytical precision.

### *1.2 AI Power of Large Language Models*

The AI boom ignited by ChatGPT in 2022 has led to breakthrough advances in Large Language Models (LLMs) across natural language understanding and generation, multimodal information processing, and few-shot learning [20-22]. With emerging capabilities in language comprehension, contextual learning, multimodal reasoning, and human-like decision-making, LLMs have evolved beyond text generation engines into versatile, general-purpose problem-solving frameworks. The technical evolution of LLMs is unfolding along two distinct trajectories：The first involves “General intelligence models” that integrate multimodal understanding capabilities and universal world knowledge (such as GPT-4, Claude-3.5, and Gemini) [23, 24]. These models acquire world knowledge foundations through large-scale pre-training and demonstrate superior performance in tasks requiring cross-domain knowledge integration and multimodal information understanding. The second path, emerging since late 2024, focuses on “Reasoning-centric models” (e.g., GPT-O1, Claude-Thinking, DeepSeek-R1), which exhibit superior performance in logical deduction, complex problem solving, and structured reasoning tasks [25-27]. These reasoning-oriented models have shown particular promise in complex scenarios such as traffic crash analysis, where integrating heterogeneous data and constructing coherent causal chains is critical—especially under conditions of incompleteness or contradiction.

Large language models have demonstrated unprecedented potential in traffic accident analysis. Research indicates that LLMs can effectively integrate heterogeneous multimodal data analysis. For example, AccidentGPT reconstructs accident process videos through multimodal inputs (audio, image, video, text, spatial and/or temporal tabular data), analyzes accident causes and liability determination, addressing the subjectivity and inefficiency issues of traditional manual analysis [28]. Furthermore, large language models have shown remarkable capabilities in traffic accident reporting analysis. For instance, an LLM-based analytical framework analyzed 500 traffic accident narrative texts, identifying previously unreported factors (such as alcohol-related incidents) in historical accident data with 96% accuracy in just 7 minutes, while human experts required an average of 25 hours [18] . More importantly, the TrafficRiskGPT system, which combines chain-of-thought with causal analysis, demonstrates that this approach outperforms traditional methods and mainstream LLMs in terms of collision prediction accuracy, reasoning speed, and overall risk evaluation metrics [29].

Despite strong language understanding and reasoning capabilities, current LLMs still have limitation in “hallucination”, meaning that they can generate fluent statements that are not supported by the provided evidence or ground truth. For example, inventing facts, misrepresenting input records, or asserting causal relationships that were never observed. Empirical evaluations show that such errors persist. A large human-annotated study of LLM-generated clinical notes reported a 1.47% sentence-level hallucination rate, and further categorized a subset as potentially major errors for downstream decision-making [30]. Hallucinations can be amplified by imperfect or contaminated inputs—when a single fabricated detail was inserted into otherwise realistic case vignettes, multiple LLMs repeated or elaborated on that false detail in 50–82% of outputs, and prompt-based mitigation reduced but did not eliminate this vulnerability [31]. Mechanistically, recent work argues that hallucination is not merely an occasional bug but is linked to structural properties of transformer-based generation, including a generalization–hallucination trade-off and limited self-correction reliability [32]. In summary, there is growing consensus that LLM hallucinations are more likely when prompts lack adequate contextual grounding and explicit constraints, and can thus be mitigated through carefully designed prompts.

#### 1.2.1 LLM agent limitations

In recent years, AI agents have found widespread applications across various domains, including scientific research, healthcare, and engineering. While the definition of an AI agent remains broad and evolving [33-36], there is growing consensus that LLM-based agents—autonomous systems powered by large language models—are capable of performing complex tasks with minimal human intervention. These agents leverage the reasoning, planning, and generative capabilities of foundation models to act in dynamic and uncertain environments. However, single-agent systems suffer from several inherent limitations.

First, the world knowledge boundaries and internal thinking patterns of single agents are constrained by the capabilities of the underlying LLM itself. Current research has shown that different types of large language models exhibit significant differences in performance when executing specific tasks [18, 37, 38].

Second, complex tasks often require step-by-step decomposition, and a single agent struggles to process multi-dimensional and multi-modal information streams simultaneously. Research indicates that single-agent systems tend to underperform across all subtasks when handling multi-modal data, particularly when the data is incomplete or contains contradictions [39, 40].

Third, the degree of specialization limits the ability of a single agent to achieve expert-level performance across multiple domains. For instance, a single agent often fails to comprehensively address the diverse subdomains involved in traffic crash analysis, especially in cross-disciplinary tasks spanning EDR data, visual evidence, and legal interpretation—leading to shallow analysis and reasoning bias [41, 42].

Furthermore, the inevitable hallucination problem of LLMs can lead to the propagation of errors and accumulation of risk. Mistakes made in early stages may compromise all subsequent steps, and a single agent typically lacks effective mechanisms for error detection and correction [43].

In summary, single-agent systems remain fundamentally constrained in knowledge coverage, task decomposition, domain specialization, and error robustness—highlighting the urgent need for more collaborative, modular AI architectures.

#### 1.2.2 Multi-agent collaboration

To address the inherent limitations of single-agent systems, the research paradigm is shifting toward multi-agent collaboration. The core principle of multi-agent systems is "divide and conquer," whereby tasks are decomposed into subtasks (or systems are modularized), and specific agents are designed for each subtask [40, 41] . Furthermore, agents must be tailored to the requirements of subtasks based on the characteristics of LLMs [44] . For instance, tasks involving image recognition necessitate agents built on vision-based large models, whereas tasks requiring logical reasoning are best handled by models equipped with chain-of-thought capabilities.

Recent research has demonstrated the superior performance of multi-agent collaboration across diverse application domains. In autonomous driving scenarios, multi-agent systems that integrate reinforcement learning, large language models, and retrieval-augmented generation (RAG) mechanisms not only enhance human-like behavior and interpretability but also outperform alternative reinforcement learning approaches on critical metrics including merge success rates and collision rates [45]. In traffic flow prediction, xTP-LLM, which converts multimodal traffic data into natural language descriptions, exhibits significantly higher accuracy compared to deep learning baselines while providing intuitive and reliable explanations for traffic predictions [46] . Furthermore, regarding interpretability, researchers have proposed a closed-loop multi-agent collaborative system capable of automatically completing the entire process from model conception to code implementation, evaluation, and iterative optimization. This approach not only achieves substantial improvements in accuracy and robustness on classical models such as IDM, MOBIL, and LWR but also maintains model interpretability and demonstrates strong generalization capabilities across datasets [47].

### *1.3 Research Objectives and Contributions*

To address the challenges in traffic crash analysis, this study develops an AI-driven multi-agent collaborative framework to investigate the pre-crash phase preceding the first collision event. The primary objective is to construct and validate a two-stage AI framework that reduces reliance on human factors when analyzing fragmented, multimodal crash data, thereby supporting expert investigations of pre-crash causal mechanisms more efficiently. Specifically, the proposed AI-agent framework performs the following phases: (Phase I) interprets crash scenarios from multimodal data sources and produces a high-level narrative of the crash event; and (Phase II) conducts temporal reasoning to identify the first collision event and its corresponding pre-crash event Data Recorder (EDR) record(s), together with an interpretable rationale for the inference.

The main contributions of this study are as follows:

a) We propose a two-stage multi-agent architecture that synthesizes fragmented multi-modal inputs to reconstruct crash scenarios and identify first collision events. Validated on 277 real-world cases, the framework demonstrates robustness in handling potential incomplete fields and conflicting information.

b) Domain-specific reasoning anchors and structured prompts promote consistent results across different LLMs, with three models achieving identical outcomes, suggesting that performance derives from prompt engineering rather than model-specific characteristics.

This study conducts a zero-shot evaluation of the LLMs' capabilities, which means that the employed initial LLMs are used without any domain-specific training, fine-tuning, or learning from datasets. This design demonstrates that our AI-Agent framework's effectiveness derives from the multi-agent collaborative architecture and prompt engineering design rather than from memorization of specific crash patterns. Consequently, the proposed framework enables researchers and industry practitioners to address domain problems while reducing the cost of collecting training labels, maintaining fine-tuned models, or running model retraining pipelines.

Overall, this research presents a multi-agent, prompt-structured analytical workflow for combining crash narratives, scene diagrams, and EDR data to support pre-crash analysis and first-collision identification.

## 2 METHODS

This section describes the proposed AI-driven multi-agent framework, including the data processing, the design of individual agent components, and the overall workflow construction. Furthermore, this section details the evaluation of the framework in identifying first collision events.

### *2.1 AI-driven Multi-agent Analytical framework*

The AI-driven multi-agent analytical framework proposed in this research operates through a structured two-phase workflow. The framework's architecture (shown in **Figure 1**), demonstrates how multiple specialized agents work in concert to transform raw, heterogeneous crash data into actionable insights about pre-crash vehicle behaviors and collision dynamics.

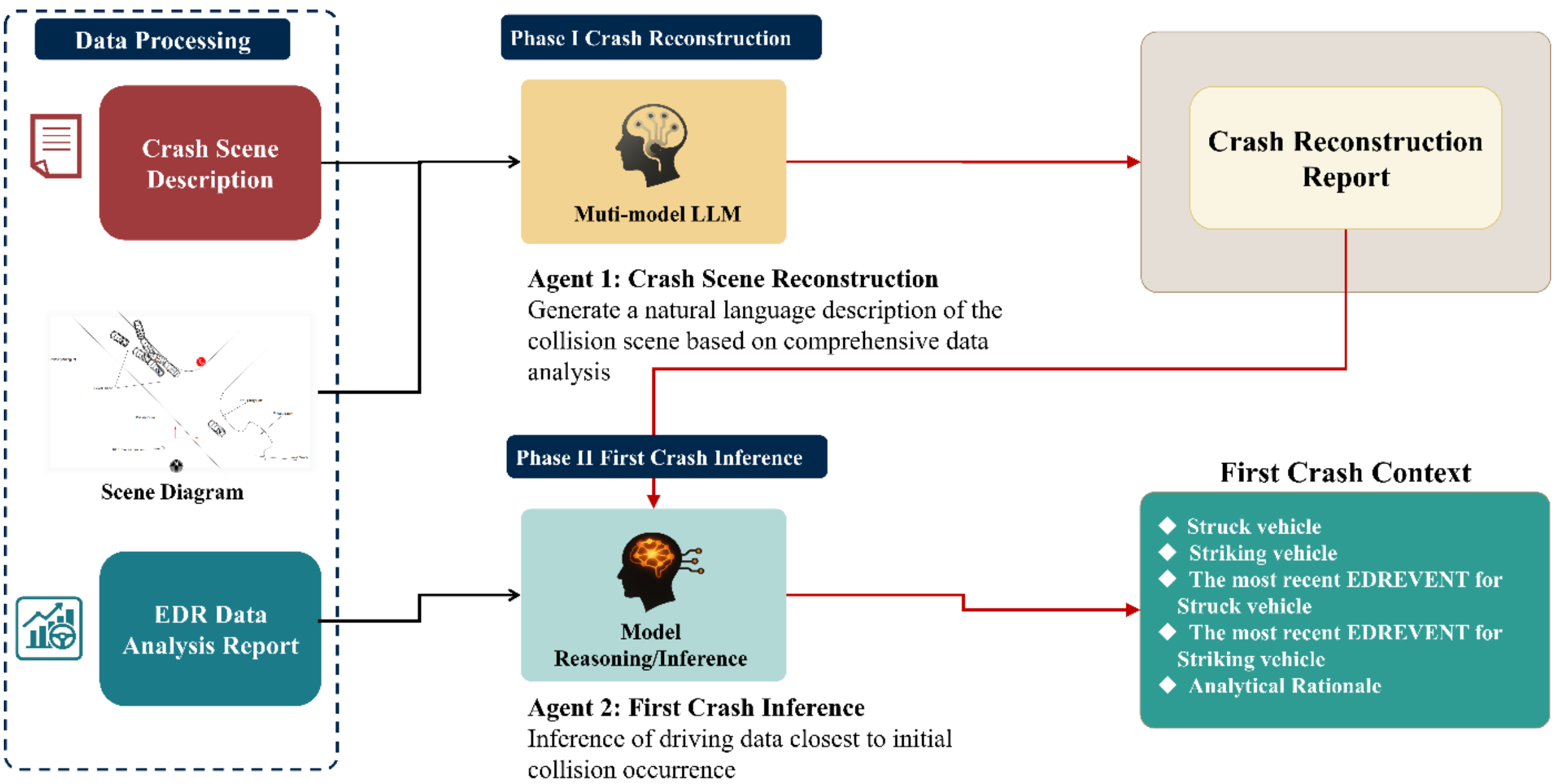

**Figure 1. Overview of Proposed AI Agent Analytical framework.**

**Step1:** The raw database data are transformed into structured natural language inputs through predefined rule-based coding. This process generates three key components: the **Crash Scene Description** containing detailed textual narratives of the collision event, the **EDR Data Analysis Report** presenting vehicle dynamics data in an interpretable format, and the **Scene Diagram** providing visual representation of the crash scenario. These processed outputs serve as standardized inputs for the subsequent agent-based analysis phases.

**Step 2:** The Crash Scene Description and Scene Diagram are fed into the Phase I Agent. This agent comprehensively understands the multi-dimensional crash data by processing all input data, capturing the essential elements of the crash scene through a predefined internal reasoning process, and leveraging multi-source data for assistance and calibration. The output is a structured, complete reconstruction of the crash scene.

**Step 3:** The output from the Phase I Agent, along with the **EDR Data Analysis Report**, is input into the Phase II Agent. This agent focuses on initial collision inference, a critical task for identifying the most relevant EDR events and distinguishing between the striking and struck vehicles. The Phase II Agent employs reasoning mechanisms to analyze the temporal patterns of EDR data within the reconstructed crash scene. By cross-referencing the reconstructed narrative with the time series of EDR events, the agent can infer the specific EDR record most relevant to the initial collision, even in complex scenarios involving multiple recorded events or ambiguous vehicle roles.

This AI analysis framework involves multi-agent collaboration, with each phase building upon and validating the previous one. The reconstruction from the first phase provides essential contextual grounding for the inference operations in the second phase, while the EDR analysis in the second phase reveals insights that deepen the understanding of the initial reconstruction. This iterative refinement process enables the system to handle majority scenarios with incomplete data, conflicting information, or multiple plausible interpretations. The final output includes identification of vehicle roles, determination of the most relevant EDR event for each vehicle involved, and an interpretable rationale for the inference.

### *2.2 Data Processing*

#### 2.2.1 Data Acquisition

In this study, we use the Crash Investigation Sampling System (CISS) dataset maintained by the National Highway Traffic Safety Administration (NHTSA) as the source of evaluation cases. CISS records detailed information from real-world motor vehicle crashes in the United States, including crash narratives in text form, structured tabular data (vehicle dynamics parameters, environmental conditions, road configurations, vehicle attributes), and scene diagrams from crash sites. CISS was selected as the data source based on two key characteristics. First, the data originates from field investigations of actual crash events rather than simulated or experimental conditions. Second, the dataset contains temporal records from EDRs. EDRs capture vehicle states during approximately the 5 to 10 seconds preceding crash trigger events, recording variables including longitudinal velocity, brake pedal position, accelerator pedal position, and lateral acceleration. These time-stamped continuous measurements provide objective evidence of vehicle behavior evolution before collisions.

This study utilized 277 cases from lead vehicle deceleration (LVD) crashes occurring between 2017 and 2022 to construct a database for evaluation of the proposed AI-agent framework.

#### 2.2.2 Transforming Unstructured Crash data into Structured Natural Language Format

To enable unified processing of heterogeneous multi-source information within the multi-agent framework, raw multimodal crash data were systematically transformed into a consistent structured natural language format (shown in **Figure 2**). The purpose of this transformation was to establish explicit contextual relationships for each data element. Research indicates that LLM reasoning performance depends heavily on input context completeness; direct input of context-free data often leads to “hallucinations” or “logical drift” during model reasoning due to missing contextual meaning [**48-50**]. Specifically, we established clear associations between numerical values, coded variables, and their corresponding vehicle entities, environment conditions, temporal references, and crash events, thereby reducing the risk of model misinterpretation from the input stage.

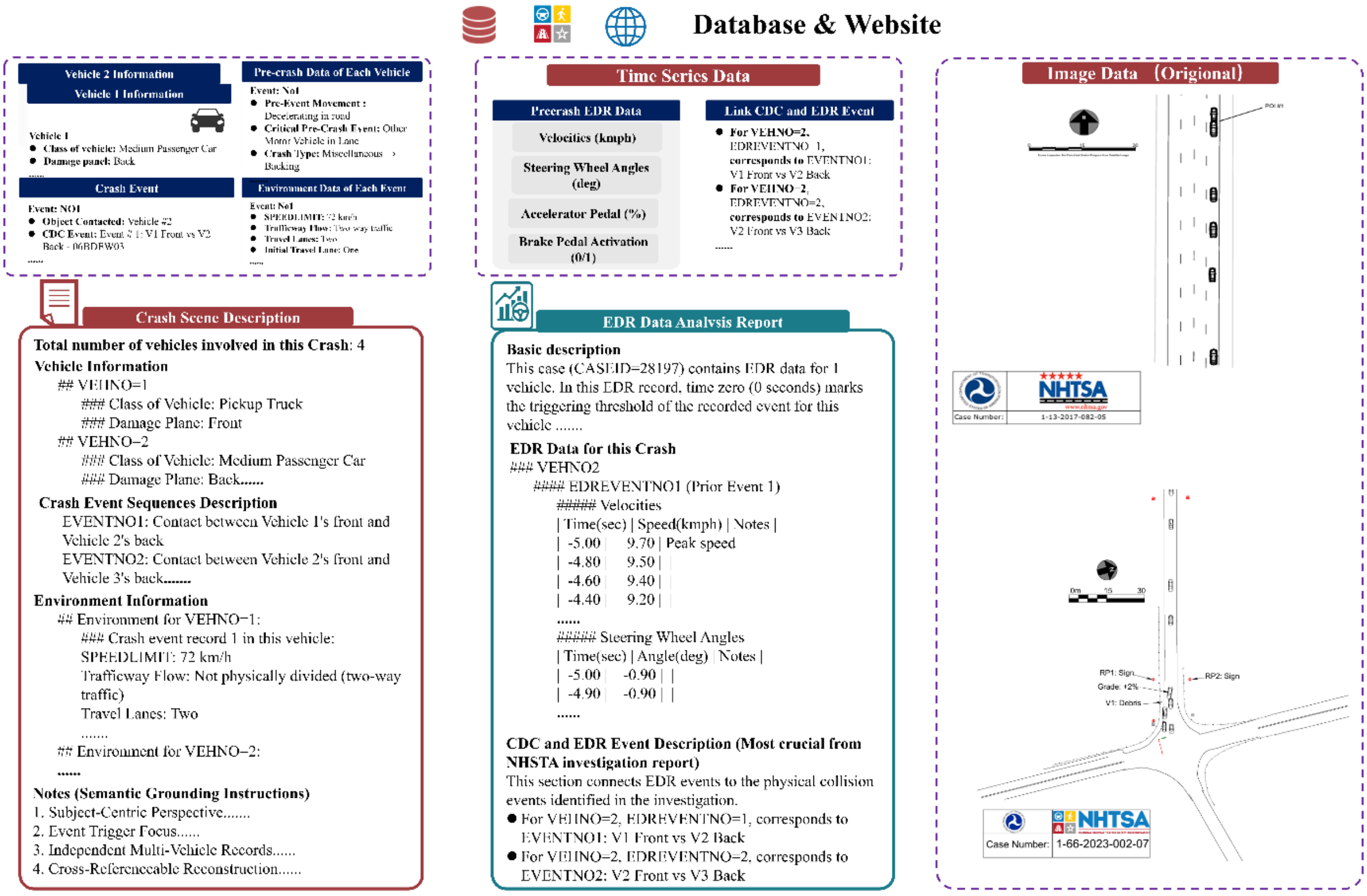

**Figure 2. Unstructured data to natural language text conversion process. Raw multimodal data is transformed into natural language text with explicit context to support semantic parsing and reasoning by large language models.**

**Textualization of crash information.** Key information related to crash scenarios (vehicle types and critical features, damage patterns, road and environmental factors, collision event descriptions) was organized into concise natural language summaries. Summaries were structured separately by vehicle, clearly distinguishing participating vehicles and their relative relationships, summarizing key collision processes, and preserving road and environmental conditions with discriminative significance for crash formation. For brevity, the output of this step is referred to as "**Crash Scene Description**" and serves as an input to the framework.

**Textualization of temporal data.** Raw temporal data from EDRs were preprocessed into structured textual records that preserve the original record sequence of the vehicle in the CISS dataset. This alignment ensured that "pre-crash behavior evolution" could be consistently mapped in subsequent reasoning stages, avoiding temporal reference deviations when multiple EDR segments or unclear event boundaries existed. For brevity, the output of this step is referred to as "EDR Data Report" and serves as an input to the framework.

**Preservation of Image data in original format.** Scene Diagrams were maintained in original format for direct input to vision-capable models. Scene diagrams in CISS were created by National Center for Statistics and Analysis (NCSA) investigation teams based on field measurements and interview records. The images contain vehicle positions, road geometry, scale annotations, and distance references [51].

This transformed data possessed a unified textual format (markdown format), with each information element embedded with necessary context.

## *2.3 Phase I Agent Design: Crash Reconstruction*

This framework divides precrash analysis into two phases, each executed by different types of large language models. Current models face challenges in simultaneously optimizing both multimodal understanding and long-chain reasoning. The framework therefore assigns collision scene reconstruction to general-purpose multimodal models (Phase I) and first crash inference to reasoning-centric models (Phase II). Phase I output serves as an information bridge, converting multimodal information from images and text into unified text format. This allows Phase II to focus on text-based reasoning without processing images. Each phase uses the model type best suited to its task characteristics, improving overall framework performance.

In this framework, prompts serve as the only communication channel between users and LLMs. Models generate responses based solely on prompt content without executing external code or accessing databases.

#### 2.3.1 Task definition

The Phase I agent takes two inputs: the scene diagram (image) and the crash scene description (text). It produces a structured Accident Reconstruction Report that is used as the input to Phase II.

This task requires reliable interpretation of vehicle spatial relationships in the scene diagram and consistent integration with text evidence. We therefore use a general-purpose multimodal LLM. We screened several mainstream models using CISS scene diagrams. These diagrams are stored as PDFs and often include distracting elements such as logos, tables, scales, compass markers, and reference-point labels. Most models tended to extract visible text under this interference rather than interpret vehicle positions, including GPT-4, GPT-4O, Grok3, Deepseek-v3, Claude, and Gemini Pro [24, 52, 53]. We found that only Claude 3.7 extracted crash-relevant spatial information more consistently without additional prompt tuning, so we selected it as the Phase I model (as of July 2025).

#### 2.3.2 Phase I Prompt Construction

**Figure 3** shows the complete prompt structure for Phase I Agent. All prompt content (role definition, data processing rules, analysis steps, and output format) is written by researchers as instructions to the model. Each component is explained below.

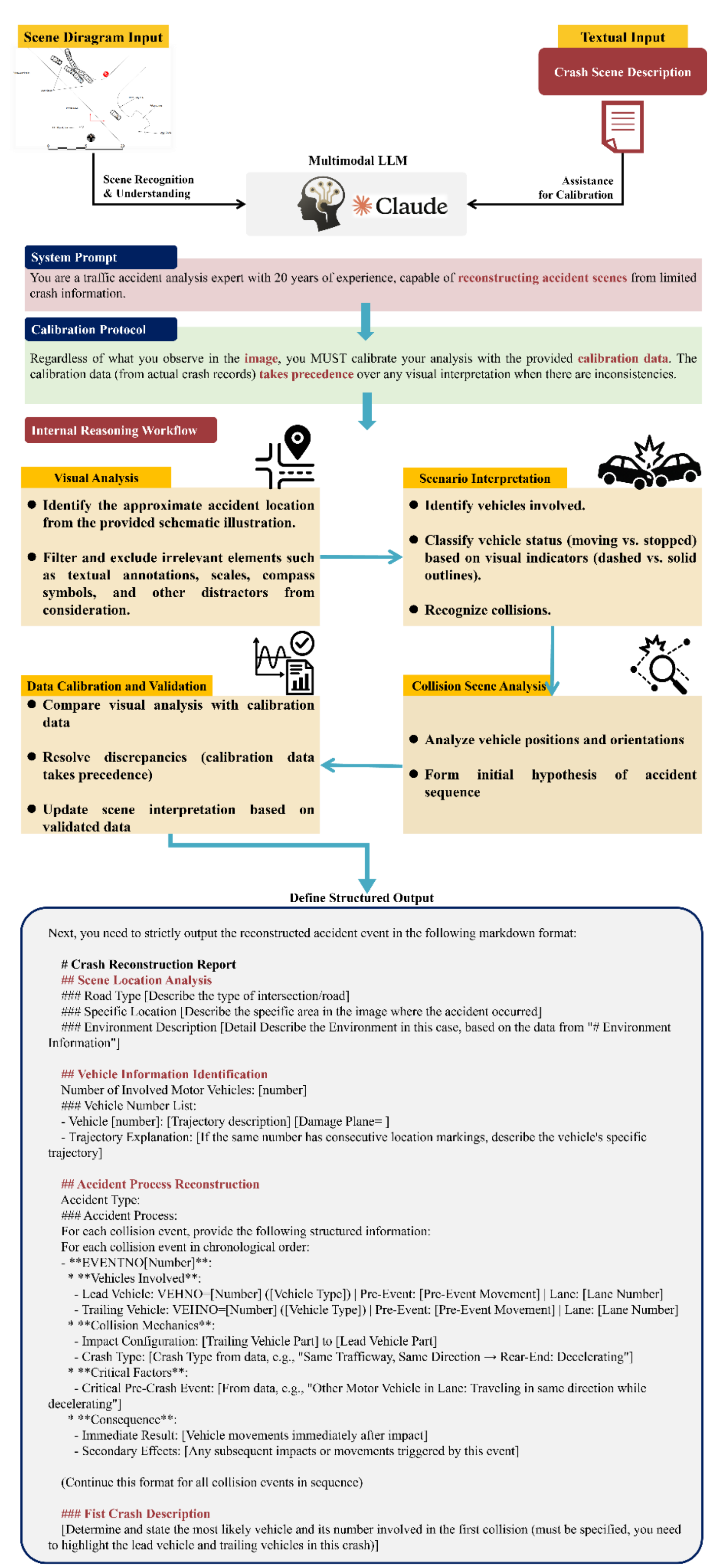

**Figure 3. Prompt design for accident-reconstruction agent.**

#### *2.3.2.1 Role Definition (System Prompt)*

Phase I agent is defined as a traffic accident analysis expert with 20 years of experience, capable of reconstructing accident scenes under limited information. This guides the model to organize information from an accident analysis perspective rather than arbitrary narratives.

#### *2.3.2.2 Evidence Priority Rules (Calibration Protocol)*

The calibration protocol instructs the model to use both the image and the text-based crash records, and to treat the text-based calibration data as higher priority when conflicts occur. This reduces the impact of diagram noise and visual ambiguity.

#### *2.3.2.3 Internal Reasoning Workflow*

The prompt also specifies a four-step workflow that matches the blue-arrow flow in Figure 3. These steps are written in the prompt and are not inferred after running the model. The bullets in the black and red boxes summarize the key instructions. The model first performs visual analysis to extract vehicle identifiers, motion states, and likely collision points while filtering out common distractors. It then interprets the scenario by confirming involved vehicles and their spatial relationships. Next, it analyzes the collision scene to construct an initial event sequence with emphasis on the first collision. Finally, it calibrates the reconstruction result against the crash records and revises the results when inconsistencies are found.

#### *2.3.2.4 Structured Output Format*

This study designed a structured output format for Phase I agent to compress collision event content. Output includes vehicle damage information, chronological description of all collision events, and preliminary positioning of the first harmful event.

Predefined format serves two purposes. First, it constrains the model to output standardized reports, reducing omissions from free-text generation. Second, it enables Phase II to directly read key fields (VEHNO, event sequence, first harmful event description) as stable input for subsequent reasoning.

## *2.4 Phase II Agent Design: First Crash Inference*

### 2.4.1 Task definition

Phase II focuses on logical reasoning under complex conditions rather than multimodal understanding. EDR systems may record multiple events, some unrelated to the target collision. Records across vehicles may be asynchronous, and some vehicles may have no usable EDR data. This requires temporal analysis and causal inference under multiple constraints.

Phase II therefore uses reasoning-specialized models, including **DeepSeek-R1, Grok 3-mini and Gemini 2.5 Pro** [25-27]. These models excel at chain-of-thought reasoning, performing multi-step logical deduction before providing answers, suitable for complex conditional judgment tasks. Unlike Phase I multimodal models, Phase II requires logical reasoning depth rather than image understanding capabilities.

### 2.4.2 Phase II Prompt Construction

**Figure 4** shows Phase II Agent prompt structure. Each component is explained below.

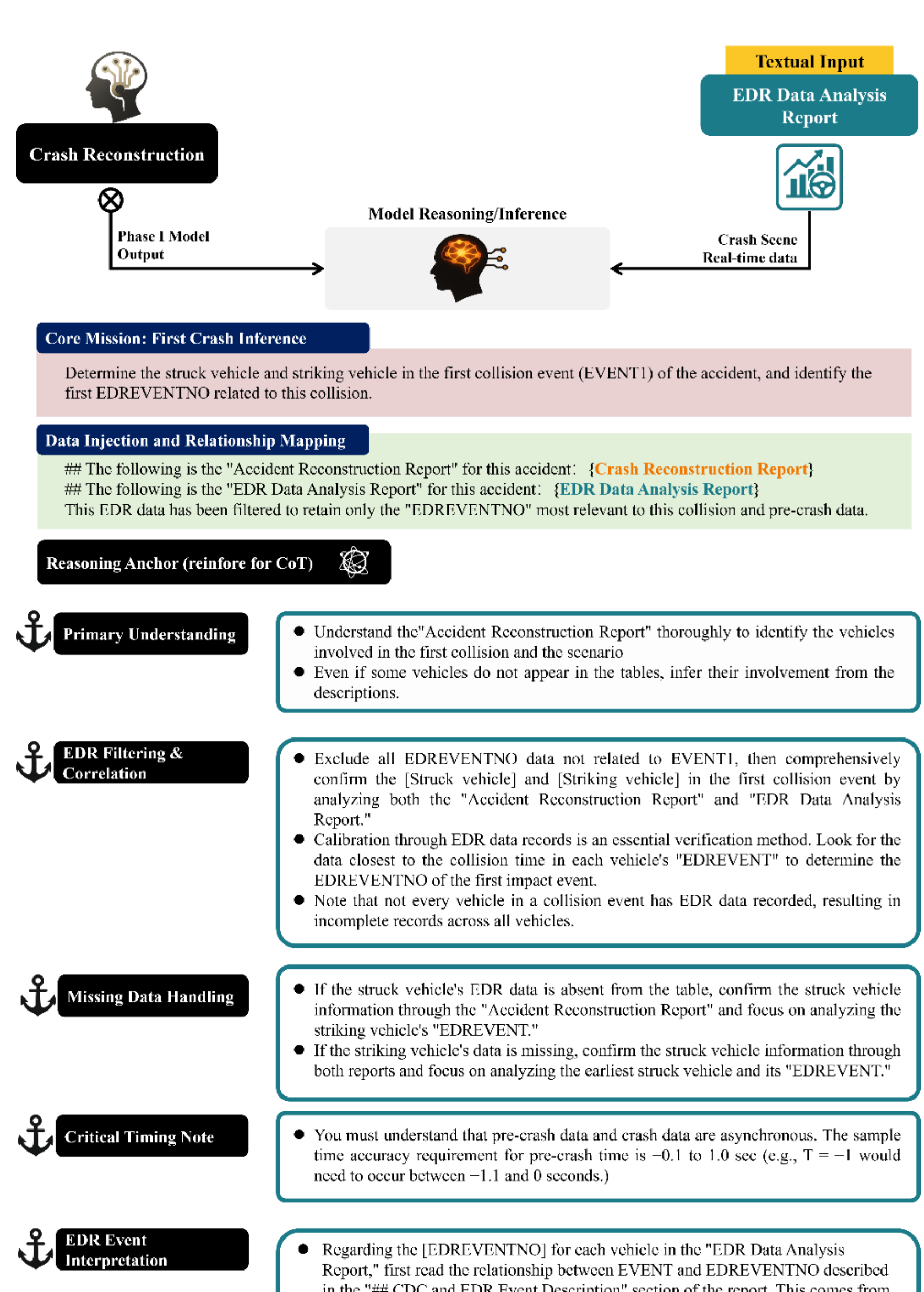

**Figure 4. Prompt design for crash inference agent.**

### *2.4.2.1 **Mission and Inputs***

This prompt declares the core task: combine accident reconstruction report and EDR data report to determine striking and struck vehicles in the first harmful event, and identify most relevant EDR record numbers for each.

The prompt embeds two complete datasets as reasoning basis: Phase I accident reconstruction report and preprocessed EDR Data Analysis Report (including temporal records for each vehicle and EDR event to collision event correspondence). Embedding all required information in the prompt ensures the model obtains complete background in a single reasoning session.

### *2.4.2.2 **Reasoning Anchors***

To reduce reasoning drift under various EDR conditions and avoid misuse of noise records, we embedded five structured reasoning rules in the prompt. These anchors are not post-hoc explanations of model behavior, but constraints the model must follow sequentially.

**Anchor 1: Primary Understanding**. Before analyzing EDR, the model must fully understand Phase I reconstruction report, clearly identifying vehicles involved in first harmful event and their scene relationships. This step prevents context-free data matching.

**Anchor 2: EDR Filtering & Correlation**. The prompt defines mandatory filtering rules to remove records that are unrelated to or unreliable for first harmful event (e.g., mappings marked as "unknown event/unrelated event", records claiming existence but lacking actual data tables). After filtering, the model cross-confirms vehicle roles using reports from both vehicles and selects event numbers best correspond to the first harmful event.

**Anchor 3: Missing Data Handling**. When one party lacks EDR records, the model falls back to reconstruction report to confirm vehicle role and marks "No Record" in the output. The focus then shifts to analyzing the other party's valid EDR evidence, avoiding fabricating event numbers or abandoning inference.

**Anchor 4: Critical Timing Note**. The prompt states that EDR time zero (T=0) is the trigger moment, not necessarily the actual collision instant. The two may deviate due to trigger thresholds and sampling mechanisms. The model performs time matching within a given tolerance range, avoiding event mismatches from misunderstanding the time baseline.

**Anchor 5: EDREVENTNO Interpretation**. When a single physical collision triggers multiple EDR records (e.g., the same EVENT1 corresponding to multiple EDREVENTNOs) with overlapping time periods, the model evaluates temporal proximity and dynamic change characteristics to select the record most closely aligned with the first harmful event, providing justification for its choice.

These reasoning anchors encode key EDR analytical judgment rules into model-executable textual instructions, enabling different reasoning models to converge on consistent inference conclusions under the same input.

### *2.4.2.3 **Structured Output with Rational***

The structured output format serves two purposes. First, standardized output enables automated evaluation through direct field-level comparison with manually annotated labels. Second, requiring the model to provide its analytical basis ensures interpretability, allowing researchers and practitioners to review the model's inference reasoning and judge the conclusion validity.

Phase II Agent prompt encodes professional EDR data analysis judgment rules into reasoning anchors embedded in a structured prompt (Figure 4). This enables reasoning-specialized models to perform accurate temporal reasoning and causal inference without domain-specific training while producing interpretable and verifiable analysis results for validation.

### *2.5 Performance Evaluation of LLM*

The evaluation dataset for the AI-Agent Framework consisted of 277 LVD cases. Each case included at least one involved vehicle with an available EDR record. Among them, 238 cases were classified as **Simple EDR cases**, which exhibited a one-to-one relationship between crash events and EDR records; that is, every crash event for any vehicle corresponded to a single, unique EDR record. Because the event–record mapping was unambiguous in these cases, we implemented a rule-based evaluation script to directly compare the AI-Agent outputs with the case records and compute agreement on the predefined output fields.

The remaining 39 cases were classified as **Complicated EDR cases**, characterized by a many-to-one relationship in which a single crash event for a vehicle corresponded to multiple EDR records. These records could also overlap in time and might contain inconsistencies (see Appendix for representative EDR examples). Notably, the terms ***Simple*** and ***Complicated*** are used only to distinguish the complexity of the EDR event mapping and do not imply differences in crash severity or other case attributes.

For the 39 Complicated EDR cases, two human researchers independently reviewed each case. While both were researchers with over three years of research experience in traffic safety and driver behavior, neither had formal training in accident reconstruction. Using the CISS Case Viewer (**https://crashviewer.nhtsa.dot.gov/ciss**), they accessed the full case materials (e.g., images, crash narratives, and complete EDR files) and documented their first-crash judgments and supporting evidence in a structured text format. Each researcher completed the review of all 39 cases within a two-week window, recording the analysis time for each case.

To reduce omissions and enhance consistency, we employed a third-party large language model (GPT-o1) as an auxiliary consistency-check tool. It is important to note that this model was not part of the proposed AI-Agent framework and did not make final determinations. Its role was strictly limited to flagging potential conflicts or overlooked evidence (e.g., timeline inconsistencies or abnormal record coverage), prompting the researchers to re-verify against the original materials. Final reference labels for the Complicated EDR cases were established through evidence-based adjudication by the two researchers. We also quantified the deviation between each researcher’s initial independent conclusion and the final consensus label.

Once the ground truth was established, the AI-Agent framework was formally evaluated using an output structure consistent with the human analysis records. The model's output was compared against these reference results to quantify its inference performance.

## 3 RESULTS

### *3.1 Overall Performance*

#### 3.1.1 AI-driven framework performance

The comprehensive evaluation of the AI-driven multi-agent analytical framework was conducted across all 277 LVD cases from the CISS dataset. A total of **4,155** experimental trials were conducted, with each of the 277 cases evaluated five times across three reasoning LLMs (277 × 5 × 3), comprising 1,190 trials for the 238 Simple EDR cases and 195 trials for the 39 Complicated EDR cases per model (Total of Simple EDR cases = 1190 * 3 LLMs, Complicated EDR cases = 195 * 3 LLMs). The evaluation criterion was defined—for each experimental trial, our framework was required to infer four critical outputs: (1) the striking vehicle, (2) the struck vehicle, (3) the most relevant EDR event for the striking vehicle, and (4) the most relevant EDR event for the struck vehicle. A single error in any of these four determinations resulted in the entire trial being classified as a failure.

The AI-driven framework achieved perfect classification accuracy across all experimental trials. As shown in **Figure 5 (a)**, the confusion matrices demonstrate that the system correctly identified all 1,190 instances in simple EDR cases and all 195 instances in Complicated EDR cases, with no false positives or false negatives recorded. This resulted in precision, recall, and F1 scores of 1.00 for both case categories.

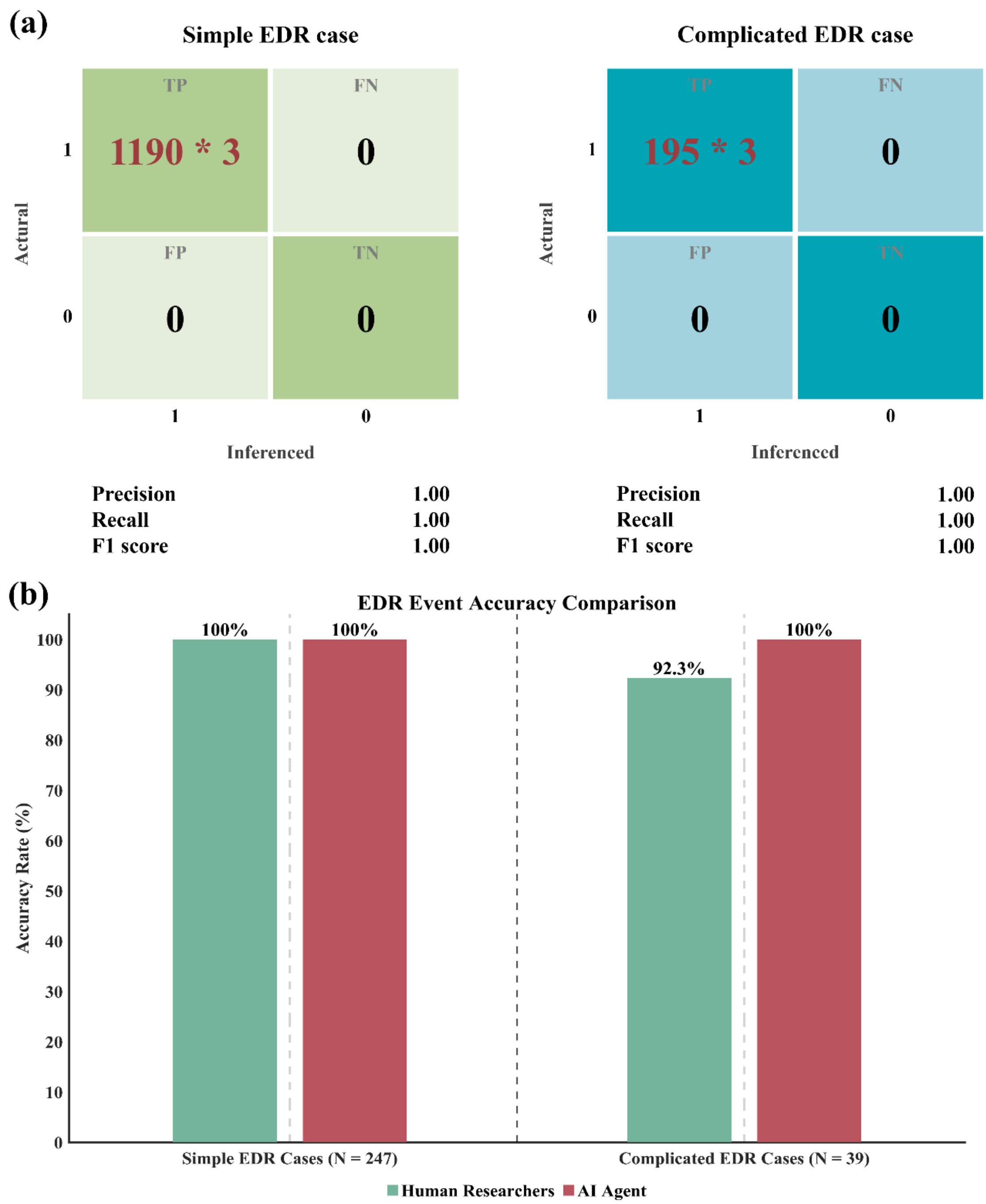

**Figure 5. Evaluation results of overall performance.**

### 3.1.2 <u>Human vs AI analytical performance</u>

A comparative evaluation showed that the research analysts achieved differential performance based on case complexity. The evaluation revealed a clear performance distinction between simple and complex EDR scenarios. In the 238 simple cases, which feature a clear one-to-one relationship between events and records, both the AI system and the two research analysts achieved 100% accuracy. This confirms that such straightforward cases are unambiguous for both automated and human review.

Performance on the 39 complex EDR cases—characterized by many-to-one relationships and occasional data errors—provides a more nuanced insight. The two individuals performing the human analysis in this study were research analysts, who are experts in crash data analysis, but not certified crash reconstruction experts. In these challenging cases, their combined accuracy was 92.31% (72 correct determinations out of 78, as shown in **Figure 5 (b)**). The performance of the research analysts could therefore be viewed as a benchmark for *non-specialist* interpretation. The significant finding is not that AI outperformed humans in an absolute sense, but that it demonstrated robust performance in complex scenarios where individuals without specialized expertise are prone to error. This highlights the AI's potential to provide stable, reproducible first-collision judgments in complex scenarios, where untrained analysts are more likely to diverge due to evidence ambiguity.

The evaluation revealed a consistent performance in vehicle role identification, with both the AI and research analysts achieving 100% accuracy. The critical differentiator was the task of determining the initial collision EDR event within complex scenarios. Here, the inherent ambiguities of multi-record cases and data logging errors led to errors in human analysis. The AI framework, however, maintained high accuracy across all five trials for each of the 39 Complicated EDR cases, demonstrating its resilience to the specific complexities that impeded human analysts.

### *3.2 Cross-model Consistency of Inference*

To evaluate the robustness and reliability of the Phase II agent's reasoning capabilities, we assessed the consistency of inference results across three reasoning models: DeepSeek-R1, Grok3-Mini, and Gemini-2.5-Pro. Each model was independently applied to all 39 Complicated EDR cases, with five experimental trials conducted per case per model, resulting in a total of 585 individual inference experiments (39 cases × 3 models × 5 trials).

The cross-model evaluation results demonstrated remarkable consistency, with all three reasoning models achieving high accuracy (100%) across all experimental trials. Each model correctly processed all 195 individual trials (39 cases × 5 trials), producing identical outputs for every inference task. The inter-model agreement rate was 100%, with no discrepancies observed between models when analyzing the same crash scenarios. This consistency demonstrates the robustness of the reasoning anchor framework and prompt design in guiding reliable analytical pathways across different LLM architectures, validating that the observed performance reflects the effectiveness of the structured reasoning framework rather than model-specific characteristics.

### *3.3 Time Use for AI Agent Analytical Framework*

**Figure 6** shows the time consumption characteristics of different AI agent configurations within the analytical framework. During phase I, the Claude agent demonstrated a consistent execution time of approximately 15.10 seconds across all configurations. For phase II, notable differences emerged among the three models employed:

**a) Claude + Deepseek-R1** incurred the highest total runtime, averaging 44.18 seconds (15.10s + 29.07s), reflecting Deepseek-R1's relatively higher computational cost and broader inference time distribution. As seen in the violin plot, Deepseek-R1 exhibited substantial variability, with response times ranging from 20s to 65s.

**b) Claude + Gemini-Pro** achieved a moderate total runtime of 32.97 seconds, supported by a more stable inference distribution centered around 18s.

**c) Claude + Grok-Mini** demonstrated the lowest total runtime at 22.71 seconds, benefitting from its second-phase agent's short response time (mean = 7.60s). This efficiency is likely attributable to Grok-Mini's design as a lightweight inference model, optimized for fast reasoning under constrained computational budgets.

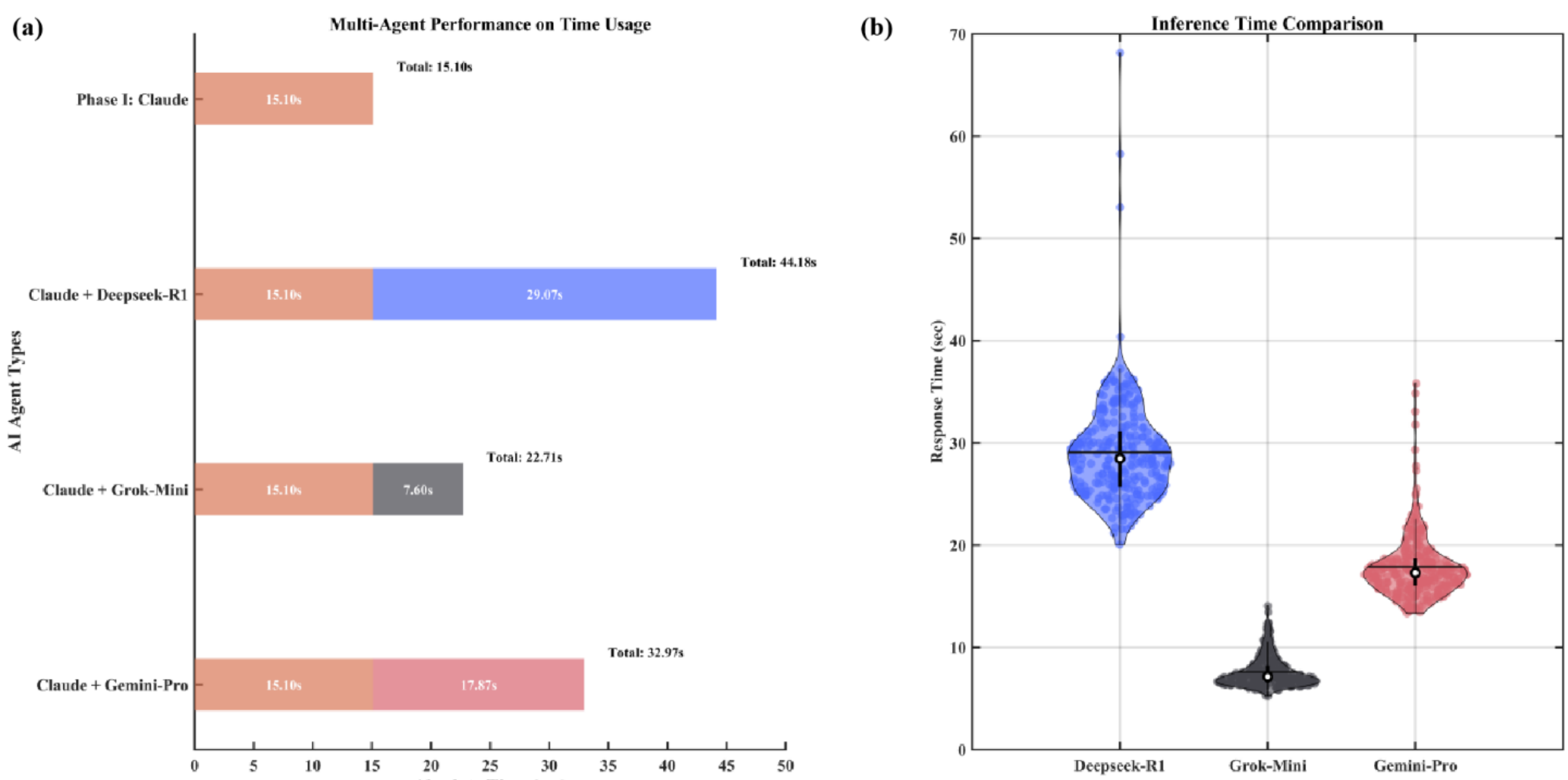

**Figure 6. Comparison of computing time across multi-agent configurations. (a) absolute execution time for Phase I and Phase II agents. (b) distribution of inference times for Phase II models.**

For comparison, human analysts required substantially longer durations to complete the same set of analytical tasks. In the 39 extreme LVD cases, two independent researchers spent **4 hours and 5 minutes** and **4 hours and 20 minutes** respectively, resulting in an average per-case processing time of **6.47 minutes**. This indicates that even the slowest AI configuration (Claude + Deepseek-R1) operated at a speed more than five times faster than human experts, while the fastest configuration (Claude + Grok-Mini) reduced analytical time by a factor of over seventeen.

These results highlight the significant time-efficiency advantage of the AI-driven multi-agent framework. By dramatically reducing analysis time without compromising result fidelity, the system demonstrates strong potential for scalable deployment in real-world traffic crash analysis tasks, particularly under time-sensitive or high-throughput conditions.

### *3.4 Quantifying the Effect of Structured Reasoning on Reasoning LLM Performance*

Although the AI-Agent demonstrated strong performance on reasoning tasks, a key question remains: to what extent does the model's performance depend on the structured reasoning guidance embedded in the prompt? To investigate this, we designed an ablation experiment on the 39 Complicated EDR cases to evaluate the contribution of Reasoning Anchors. For each case, the Phase I output was held constant as the unified input (one randomly selected result from the 15 trials per case described in Section 3.1.1). Two experimental conditions were compared:

- **With Reasoning Anchors:** The original Phase II prompt was used, containing the full Reasoning Anchors module introduced in Section 2.4.2.2, thereby guiding the model through the structured reasoning framework to interpret the context.
- **Without Reasoning Anchors:** The Reasoning Anchors module was removed, retaining only the Core Mission, Data & Relationship, and Structured Output modules, where the model relied entirely on its own capability to interpret the context.

Each case was tested 10 times independently under each condition for each of the three reasoning LLMs (Gemini 2.5 Pro, Deepseek-R1, and Grok 3 Mini), resulting in a total of 2,340 experiments (2 × 3 × 10 × 39).

Under the With Reasoning Anchors condition, the case-level accuracy was 99.7%. The models produced no errors in the Struck Vehicle, Striking Vehicle, or Struck EDR Event outputs; errors occurred only in the Striking EDR Event output. After removing Reasoning Anchors, the case-level accuracy dropped to 96.5%, a decrease of 3.2 percentage points. Although this numerical difference appears modest, the error patterns differed substantially between the two testing conditions. With reasoning anchors, errors were confined to a single output type, whereas without reasoning anchors, errors were distributed across all four output types, indicating a broader decline in reasoning stability (shown in **Fig. 7**).

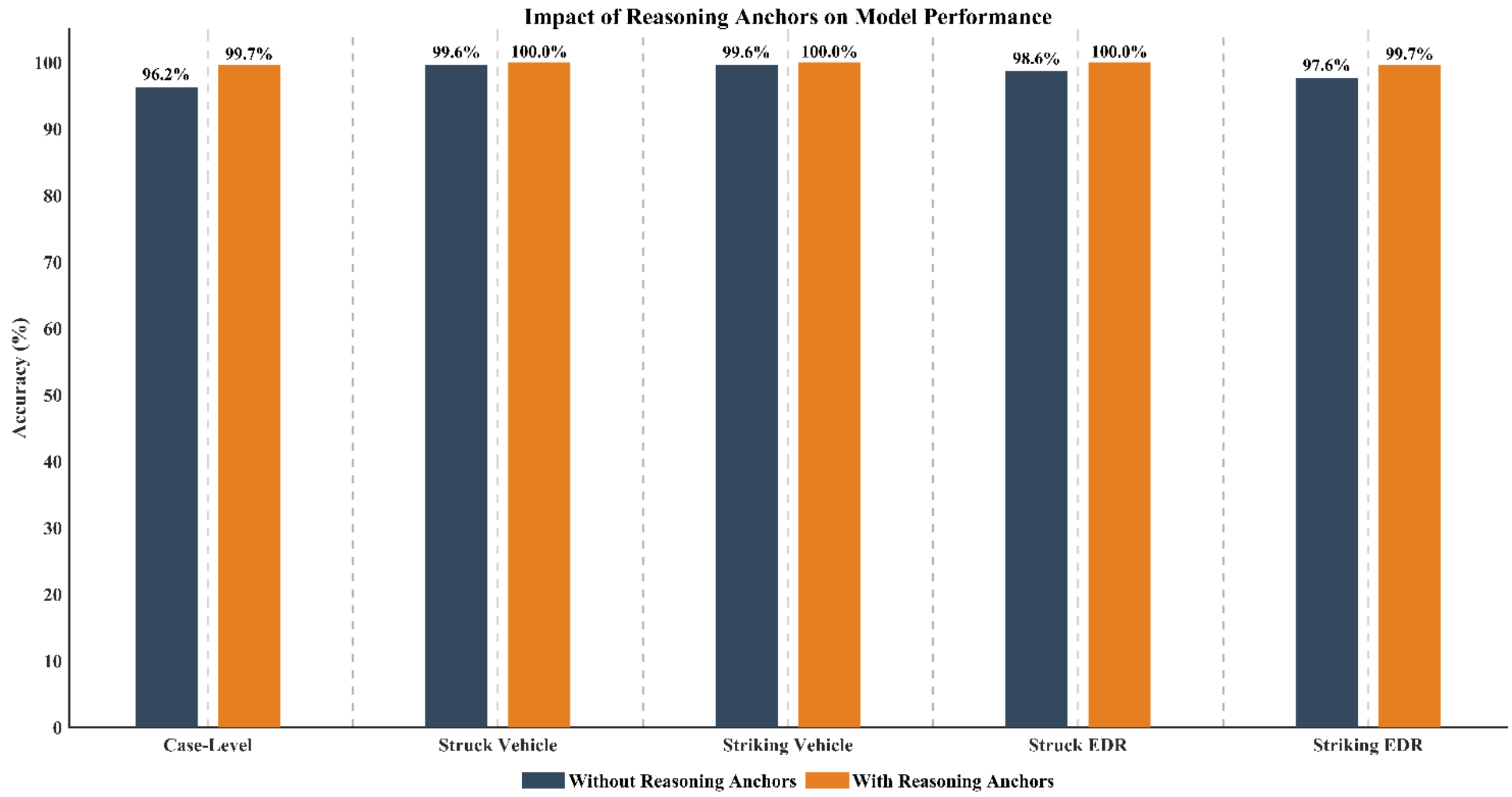

**Fig. 7. Case-level accuracy comparison between the With Reasoning Anchors and Without Reasoning Anchors conditions**

**Table 1** presents the per-item accuracy across all three LLM models for the five Complicated EDR cases that exhibited judgment errors under one or both testing conditions. Under the With Reasoning Anchors condition, performance was nearly perfect: four of the five cases achieved complete accuracy, and the remaining case (31925) showed only limited errors confined to the Striking EDR Event output, where Deepseek-R1 and Grok3-Mini each produced 2 errors out of 10 trials.

In contrast, removing the Reasoning Anchors led to substantial performance degradation in multiple cases. Accuracy declines were often pronounced (e.g., in Case 32548, dropping from 100% to 20–40% in other affected outputs), and errors were no longer confined to a single category but instead distributed across multiple output types. Across all case–model–output combinations in which differences were observed, performance without Reasoning Anchors never exceeded that under the With Reasoning Anchors condition. On average, the With Reasoning Anchors configuration outperformed the Without Reasoning Anchors configuration by 55.6%..

**Table 1. Performance of three reasoning LLMs on the five error-containing cases under the With Reasoning Anchors and Without Reasoning Anchors conditions.**

| CaseID | LLM | Output | With Reasoning Anchors (%) | Without Reasoning Anchors (%) | Diff (%) |
|---|---|---|---|---|---|
| **12988** | Gemini 2.5 Pro | Struck Vehicle | 100 | 100 | 0 |
| | | Striking Vehicle | 100 | 100 | 0 |
| | | Struck EDR Event | 100 | 100 | 0 |
| | | Striking EDR Event | 100 | **20** | **+80** |
| | Deepseek-R1 | Struck Vehicle | 100 | 100 | 0 |
| | | Striking Vehicle | 100 | 100 | 0 |
| | | Struck EDR Event | 100 | 100 | 0 |
| | | Striking EDR Event | 100 | 100 | 0 |
| | Grok 3 Mini | Struck Vehicle | 100 | 100 | 0 |
| | | Striking Vehicle | 100 | 100 | 0 |
| | | Struck EDR Event | 100 | 100 | 0 |
| | | Striking EDR Event | 100 | **30** | **+70** |
| **16231** | Gemini 2.5 Pro | Struck Vehicle | 100 | 100 | 0 |
| | | Striking Vehicle | 100 | 100 | 0 |
| | | Struck EDR Event | 100 | 100 | 0 |
| | | Striking EDR Event | 100 | 100 | 0 |
| | Deepseek-R1 | Struck Vehicle | 100 | 100 | 0 |
| | | Striking Vehicle | 100 | 100 | 0 |
| | | Struck EDR Event | 100 | 100 | 0 |
| | | Striking EDR Event | 100 | 100 | 0 |
| | Grok 3 Mini | Struck Vehicle | 100 | 100 | 0 |
| | | Striking Vehicle | 100 | 100 | 0 |
| | | Struck EDR Event | 100 | 100 | 0 |
| | | Striking EDR Event | 100 | **40** | **+60** |
| **21706** | Gemini 2.5 Pro | Struck Vehicle | 100 | **50** | **+50** |
| | | Striking Vehicle | 100 | **50** | **+50** |
| | | Struck EDR Event | 100 | **50** | **+50** |
| | | Striking EDR Event | 100 | 100 | 0 |
| | Deepseek-R1 | Struck Vehicle | 100 | 100 | 0 |
| | | Striking Vehicle | 100 | 100 | 0 |
| | | Struck EDR Event | 100 | 100 | 0 |
| | | Striking EDR Event | 100 | 100 | 0 |
| | Grok 3 Mini | Struck Vehicle | 100 | 100 | 0 |
| | | Striking Vehicle | 100 | 100 | 0 |
| | | Struck EDR Event | 100 | 100 | 0 |
| | | Striking EDR Event | 100 | 100 | 0 |
| **31925** | Gemini 2.5 Pro | Struck Vehicle | 100 | 100 | 0 |
| | | Striking Vehicle | 100 | 100 | 0 |
| | | Struck EDR Event | 100 | 100 | 0 |
| | | Striking EDR Event | 100 | 100 | 0 |
| | Deepseek-R1 | Struck Vehicle | 100 | 100 | 0 |
| | | Striking Vehicle | 100 | 100 | 0 |
| | | Struck EDR Event | 100 | 100 | 0 |
| | | Striking EDR Event | **80** | **50** | **+30** |
| | Grok 3 Mini | Struck Vehicle | 100 | 100 | 0 |
|  |  | Striking Vehicle | 100 | 100 | 0 |
|  |  | Struck EDR Event | 100 | 100 | 0 |
|  |  | Striking EDR Event | **80** | **80** | 0 |
| **32548** | Gemini 2.5 Pro | Struck Vehicle | 100 | 100 | 0 |
|  |  | Striking Vehicle | 100 | 100 | 0 |
|  |  | Struck EDR Event | 100 | **0** | **+100** |
|  |  | Striking EDR Event | 100 | 100 | 0 |
|  | Deepseek-R1 | Struck Vehicle | 100 | 100 | 0 |
|  |  | Striking Vehicle | 100 | 100 | 0 |
|  |  | Struck EDR Event | 100 | **90** | **+10** |
|  |  | Striking EDR Event | 100 | 100 | 0 |
|  | Grok 3 Mini | Struck Vehicle | 100 | 100 | 0 |
|  |  | Striking Vehicle | 100 | 100 | 0 |
|  |  | Struck EDR Event | 100 | 100 | 0 |
|  |  | Striking EDR Event | 100 | 100 | 0 |

**Notes: Each cell reports the percentage of correct outputs across 10 independent trials. Diff denotes the accuracy difference between the two conditions.**

Fig. 8 presents the token consumption under both conditions. The inclusion of Reasoning Anchors increased the input tokens by an average of 14.3% across the three models (DeepSeek-R1: 14.8%, Grok3 Mini: 14.5%, Gemini 2.5 Pro: 13.6%). Reasoning-stage tokens consumed during the model's internal chain-of-thought stage increased more substantially, by an average of 38.7% (DeepSeek-R1: 65.6%, Grok3 Mini: 8.6%, Gemini 2.5 Pro: 41.8%). This pattern suggests that the structured reasoning guidance in the prompt drove the models to engage in more extensive intermediate reasoning during inference. Finally, the output tokens increased by only 20.1% on average (DeepSeek-R1: 8.9%, Grok3 Mini: 25.7%, Gemini 2.5 Pro: 25.6%), reflecting the constraint imposed by the structured output format, which limited the variation in final output length regardless of the additional reasoning effort.

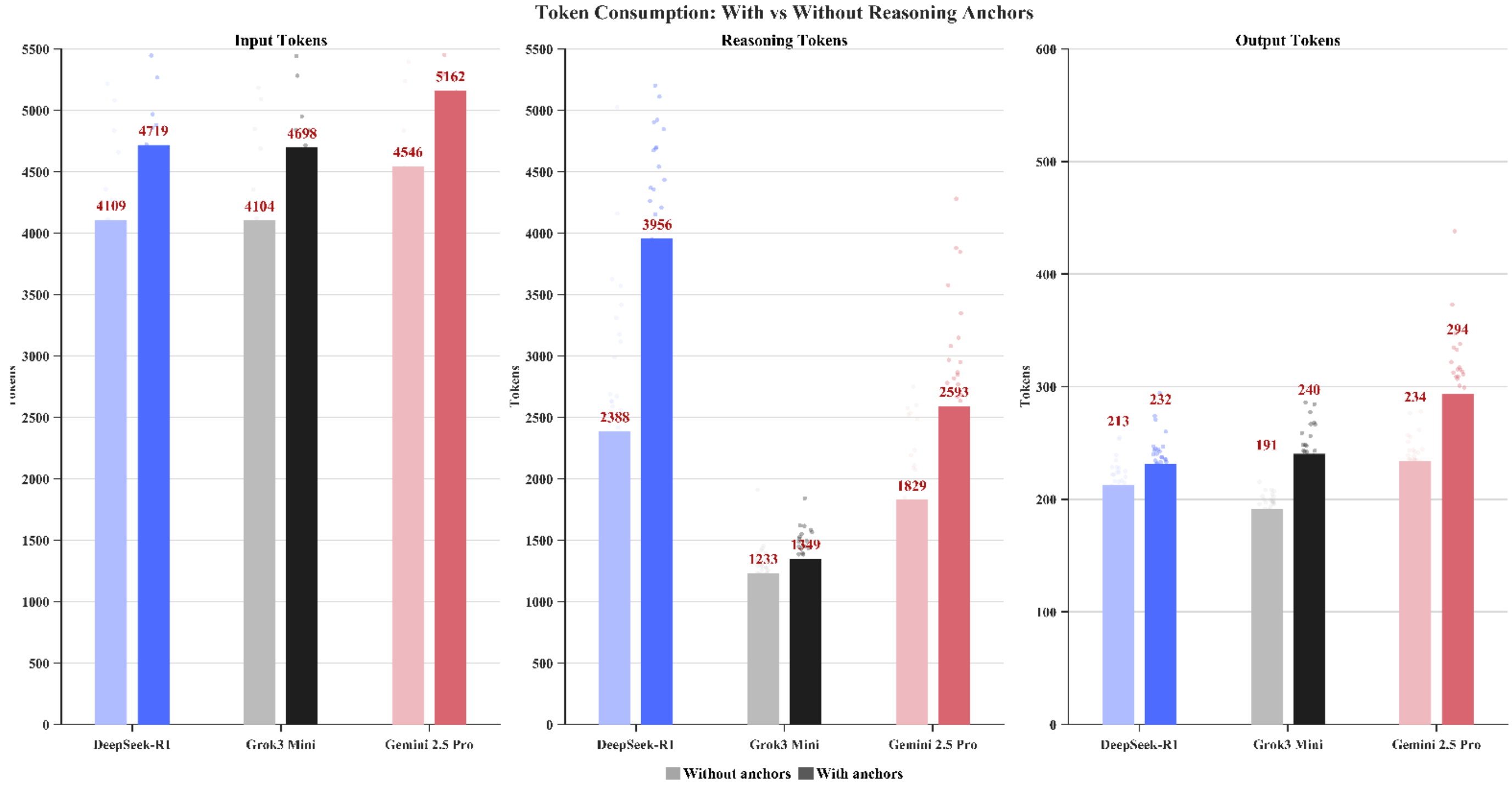

**Fig. 8 Token consumption in different conditions.**

The ablation experiment shows that removing Reasoning Anchors led to more frequent and more broadly distributed errors, especially in complex multi-vehicle collision cases. Including Reasoning Anchors mainly improved output consistency and reduced failure cases, while increasing token usage during inference.

# 4 DISCUSSION

The discussion does not aim to rank the AI system against human experts. Instead, we examine how the proposed AI-Agent framework performs in crash analysis settings that are complex and affected by uncertainty and data noise. We focus on output stability, repeatability, and practical value as decision support for non-specialists. We also examine the role of structured reasoning and its computational cost.

## *4.1 Stable Analytical Support in Various Crash Scenarios*

As reported in Section 3.1.1, we conducted 4,155 fully independent end-to-end experimental trials on 277 LVD cases, with both Phase I and Phase II agents involved in each trial. Under a fixed task definition and standardized inputs, the AI-Agent framework produced highly consistent outputs across runs. This observation does not imply that the system is error-free in all crash analysis situations. Rather, it indicates that, within the scope of the task defined in this study, the framework did not exhibit random output drift or unstable behavior across repeated executions.

In Simple EDR cases, crash events and EDR records have one-to-one mapping, making the evidence selection path relatively straightforward. Consequently, both the system and human analysts were able to reach consistent conclusions. In contrast, Complicated EDR cases present difficulties primarily at the data structure level (e.g. temporal overlaps, and potentially inconsistent records), which significantly increase the difficulty of evidence selection and logical consistency maintenance. The human researchers experiment described in **Section 2.5** showed that even research personnel with traffic safety backgrounds (nonetheless without systematic accident reconstruction training) were more prone to judgment deviations in these cases. A typical risk is that when multiple temporally overlapping records appear highly similar, analysts may be misled by database labeling errors or "seemingly reasonable" data features, deviating from the evidence combination that best align with the crash sequence [**54**, **55**]. This phenomenon is particularly evident in Case 32548 (see Appendix A).

The value of our AI-Agent framework's is not to replace expert judgment but to provide stable and reproducible analytical support. It enables non-specialist personnel to systematically conduct evidence screening and relationship determination, reduce omissions, and triage cases involving ambiguous or anomalous records. This positioning aligns with prior research emphasizing AI as a decision-support assistant to improve consistency and reduce cognitive load **[56]**. The end-to-end runtime is also practical for high-throughput workflows: on average, the framework processed each case in under one minute, compared with 6.47 minutes required for human review of the **Complicated EDR** cases. Recent research supports the efficiency of LLMs in addressing underreporting phenomena in traffic accident data, achieving 96% accuracy in processing 500 traffic accident reports in just 7 minutes **[18]**.

## *4.2 The role of structured reasoning in maintaining inference stability*

Although reasoning-oriented large language models already demonstrate strong inference capability, the ablation results in **Section 3.4** show that this capability can degrade in certain non-standard cases when structured reasoning guidance is removed. It is important to clarify that the experiments in Section 3.4 do not represent a re-evaluation of the complete AI-Agent framework. Instead, they isolate the effect of structured reasoning in Phase II by holding Phase I outputs constant and varying only the prompt structure for Phase II. The evaluation was conducted on 39 Complicated EDR cases, with each model tested ten times under each condition to assess inference stability rather than single-run performance.

When Reasoning Anchors were included, the models maintained a high case-level accuracy (99.7%), and observed errors were largely confined to a single output type (only 4 errors in one case). After removing structured reasoning guidance, overall accuracy decreased to 96.5%. More importantly, errors began to appear simultaneously across multiple output dimensions, including vehicle role identification and EDR event selection. This shift in error distribution suggests that the issue is not occasional misclassification at an isolated step, but increased instability in the reasoning process under complex conditions.

Detailed examination of the five error-prone cases further supports this interpretation. In cases with multiple time-overlapping EDR records, the absence of explicit reasoning constraints made the models more susceptible to relying on surface-level similarities rather than consistent temporal and causal relationships. For example, in Case 32548, the Struck EDR Event accuracy of Gemini 2.5 Pro dropped to 0% after removing Reasoning Anchors. In Case 21706, the same model failed simultaneously in collision relationship identification and EDR event inference. These concurrent failures indicate that, without structured guidance, the underlying reasoning process becomes less stable rather than failing at a single decision point.

Across all configurations where performance differences were observed, the condition without Reasoning Anchors never outperformed the condition with structured reasoning. This pattern was consistent across all three reasoning models evaluated. The consistency of this degradation suggests that the observed effect is not model-specific, but reflects a broader vulnerability of unconstrained chain-of-thought reasoning in environments with overlapping, inconsistent, or noisy information [27, 57].

In summary, reasoning anchors provide a robust methodological framework for building more reliable and reproducible LLM-based analysis systems. By explicitly addressing the challenges of analytical transparency and consistency, this prompt design technique paves the way for trustworthy AI integration in specialized, high-stakes vertical domains.

### *4.3 Computational cost and time-efficiency trade-offs*

The stability gains provided by structured reasoning are accompanied by additional computational cost. As shown in Section 3.4, introducing Reasoning Anchors resulted in only a modest increase in input tokens (14.3% on average), but a substantially larger increase in reasoning-stage token consumption (38.7% on average). This pattern indicates that structured reasoning does not primarily affect model behavior by supplying more input information. Instead, it encourages the model to allocate more computational resources to internal reasoning under explicit constraints.

While different models showed varying levels of increase in reasoning token usage, the overall trend was consistent. In contrast, output token growth remained relatively limited (20.1% on average), reflecting the use of a structured output format that restricts unnecessary verbosity. Importantly, even with the additional reasoning cost, the end-to-end analysis time of the AI-Agent framework remained significantly shorter than that required by human reviewers for Complicated EDR cases (AI-Agent: <1 min/case on average; human reviewers untrained in crash reconstruction: 6.47 min/case).

From a deployment perspective, these findings suggest that the computational overhead introduced by structured reasoning is manageable and does not compromise the framework's suitability for high-throughput analysis. Moreover, they suggest a practical usage strategy: lightweight reasoning configurations may be sufficient for cases with clear and unambiguous EDR mappings, while structured reasoning is particularly valuable for cases involving multiple EDR records with temporal overlap, or potential data inconsistencies. In such cases, the additional computational cost is justified by the improved stability and reduced risk of systematic inference errors.

### *4.4 Limitation and Future Works*

Although our AI-driven analytical framework demonstrated high accuracy and robustness, this study has several limitations. First, the evaluation was restricted to LVD scenarios and does not yet cover other crash types such as side impacts, head-on collisions, or large-scale multi-vehicle pileups. Therefore, the findings should not be directly generalized beyond the tested scenario. Second, the reference labels for Complicated EDR cases were established through a structured consensus process involving trained researchers rather than certified accident reconstruction experts. Although multiple verification steps were used to reduce bias, future studies could incorporate expert-labeled benchmarks to further strengthen validation.

Future research will proceed along several directions. The immediate priority is extending the framework to additional crash types and more complex multi-vehicle scenarios to assess its generalizability. This extension will also evaluate whether the current reasoning anchor design remains effective under fundamentally different collision dynamics, or whether scenario-specific reasoning constraints are needed.

Beyond scenario extension, the accurate identification of first collision events established in this study enables systematic pre-crash behavioral analysis at scale. Once the correct pre-crash temporal window is anchored for each vehicle, researchers can extract and compare driver behavioral patterns, such as braking onset timing, deceleration profiles, and speed maintenance characteristics, across large case populations. This type of large-scale behavioral analysis is currently constrained by the time cost of manual case-by-case interpretation, and our framework provides the analytical foundation to address this bottleneck.

Furthermore, the natural-language crash reconstructions produced by the framework are inherently compatible with emerging generative AI capabilities. As text-to-image and text-to-video technologies continue to advance, structured crash sequence descriptions could serve as inputs for generating visual reconstructions of collision processes. Such visual outputs would support accident investigation, driver education, and safety communication. The reliability of these downstream applications depends on the accuracy of the underlying textual reconstruction, which this study establishes and validates.

Additionally, adaptive reasoning strategies will be explored to balance inference stability with computational efficiency, supporting practical deployment in large-scale traffic crash analysis workflows.

## 5 CONCLUSION

The primary contribution of this study lies in the development of a novel AI-driven multi-agent framework for pre-crash analysis, leveraging advanced AI to enhance the accuracy and efficiency of traffic collision analysis. The following conclusions can be drawn:

(1) The framework maintains stable and reproducible outputs across repeated experiments, providing consistent analytical support for LVD crash scenarios. This stability stems from structured integration of multimodal evidence and guided reasoning processes rather than unconstrained model inference.

(2) Structured reasoning constraints suppress inference drift and error propagation in complex scenarios. Ablation experiments show that removing reasoning anchors reduced case-level accuracy from 99.7% to 96.5%, with errors spreading from a single output type to multiple analytical dimensions. This reveals that AI reliability in safety-critical applications depends on explicit reasoning frameworks rather than model capability alone.

(3) The AI-driven framework achieved higher accuracy compared to research analysts with domain expertise but no specialized accident reconstruction training (100% vs 92.31% accuracy on complex cases), demonstrating its potential as a reliable analytical assistant for pre-crash investigation.

Future research will extend the framework to additional crash types beyond LVD scenarios and incorporate expert-labeled benchmarks to further assess generalizability. In parallel, adaptive reasoning strategies will be explored to balance inference stability with computational efficiency, supporting practical deployment in large-scale traffic crash analysis workflows. Importantly, the proposed AI-driven framework established in this study serves as an essential foundation for future in-depth pre-crash analysis research.

## REFERENCES

1. WHO. 2023. *Global Status Report on Road Safety 2023.* World Health Organization (Geneva: World Health Organization). **https://www.who.int/publications/i/item/9789240086517**.
2. NHTSA. 2025. *Early Estimates of Motor Vehicle Traffic Fatalities and Fatality Rate by Sub-Categories in 2024.* U.S. Department of Transportation (1200 New Jersey Avenue SE, Washington, DC 20590 ). **https://crashstats.nhtsa.dot.gov/Api/Public/ViewPublication/813729**.
3. Scanlon, J. M., K. D. Kusano, and H. C. Gabler, *Analysis of Driver Evasive Maneuvering Prior to Intersection Crashes Using Event Data Recorders.* Traffic Inj Prev, (2015), *16 Suppl 2*, S182-9. **https://doi.org/10.1080/15389588.2015.1066500**
4. Kolla, E., V. Adamova, and P. Vertal, *Simulation-based reconstruction of traffic incidents from moving vehicle mono-camera.* Sci Justice, (2022), *62*(1), 94-109. **https://doi.org/10.1016/j.scijus.2021.11.001**
5. Doecke, S. D., M. R. J. Baldock, C. N. Kloeden, and J. K. Dutschke, *Impact speed and the risk of serious injury in vehicle crashes.* Accident Analysis & Prevention, (2020), *144*, 105629. **https://doi.org/10.1016/j.aap.2020.105629**
6. Santos, Kenny, Nuno M. Silva, João P. Dias, and Conceição Amado, *A methodology for crash investigation of motorcycle-cars collisions combining accident reconstruction, finite elements, and experimental tests.* Engineering Failure Analysis, (2023), *152*. **https://doi.org/10.1016/j.engfailanal.2023.107505**
7. Shen, Z., W. Ji, S. Yu, G. Cheng, Q. Yuan, Z. Han, H. Liu, and T. Yang, *Mapping the knowledge of traffic collision Reconstruction: A scientometric analysis in CiteSpace, VOSviewer, and SciMAT.* Sci Justice, (2023), *63*(1), 19-37. **https://doi.org/10.1016/j.scijus.2022.10.005**
8. Vida, Gábor and Árpád Török, *Expected effects of accident data recording technology evolution on the identification of accident causes and liability.* European Transport Research Review, (2023), *15*(1), 17. **https://doi.org/10.1186/s12544-023-00591-4**
9. Evtiukov, Sergei, Elena Kurakina, Valery Lukinskiy, and Aleksey Ushakov, *Methods of Accident Reconstruction and Investigation Given the Parameters of Vehicle Condition and Road Environment.* Transportation Research Procedia, (2017), *20*, 185-192. **https://doi.org/10.1016/j.trpro.2017.01.049**
10. Davis, Gary A., *Crash reconstruction and crash modification factors.* Accident Analysis & Prevention, (2014), *62*, 294-302. **https://doi.org/10.1016/j.aap.2013.09.027**
11. Deng, X., Z. Du, H. Feng, S. Wang, H. Luo, and Y. Liu, *Investigation on the Modeling and Reconstruction of Head Injury Accident Using ABAQUS/Explicit.* Bioengineering (Basel), (2022), *9*(12). **https://doi.org/10.3390/bioengineering9120723**
12. Maulina, Dewi, Diandra Yasmine Irwanda, Thahira Hanum Sekarmewangi, Komang Meydiana Hutama Putri, and Henry Otgaar, *How accurate are memories of traffic accidents? Increased false memory levels among motorcyclists when confronted with accident-related word lists.* Transportation Research Part F: Traffic Psychology and Behaviour, (2021), *80*, 275-294. **https://doi.org/10.1016/j.trf.2021.04.015**

13. Chung, Younshik and IlJoon Chang, *How accurate is accident data in road safety research? An application of vehicle black box data regarding pedestrian-to-taxi accidents in Korea.* Accident Analysis & Prevention, (2015), *84*, 1-8. **https://doi.org/10.1016/j.aap.2015.08.001**
14. Janstrup, Kira H., Kaplan Sigal, Hels Tove, Lauritsen Jens, and Carlo G. and Prato, *Understanding traffic crash under-reporting: Linking police and medical records to individual and crash characteristics.* Traffic Injury Prevention, (2016), *17*(6), 580-584. **https://doi.org/10.1080/15389588.2015.1128533**
15. Miller, Ted R, Rekaya Gibson, Eduard Zaloshnja, Lawrence J Blincoe, John Kindelberger, Alexander Strashny, Andrea Thomas, Shiu Ho, Michael Bauer, and Sarah Sperry, *Underreporting of driver alcohol involvement in United States police and hospital records: capture-recapture estimates.* Annals of Advances in Automotive Medicine/Annual Scientific Conference, (2012), *56*, 87-96. **http://www.ncbi.nlm.nih.gov/pmc/articles/PMC3503413/**.
16. Bhatti, Junaid A. and Louis-Rachid and Salmi, *Challenges in Evaluating the Decade of Action for Road Safety in Developing Countries: A Survey of Traffic Fatality Reporting Capacity in the Eastern Mediterranean Region.* Traffic Injury Prevention, (2012), *13*(4), 422-426. **https://doi.org/10.1080/15389588.2012.655431**
17. Viano, David C., *NHTSA crash investigation sampling system (CISS) is unreliable, inaccurate and does not estimate serious injury in motor vehicle crashes.* Traffic Injury Prevention, (2025), *26*(1), 61-75. **https://doi.org/10.1080/15389588.2025.2451572**
18. Arteaga, Cristian and JeeWoong Park, *A large language model framework to uncover underreporting in traffic crashes.* Journal of Safety Research, (2025), *92*, 1-13. **https://doi.org/10.1016/j.jsr.2024.11.009**
19. Adeel, Muhammad, King , Meredith, Usman , Sheikh M., and Asad J. and Khattak, *Advancing crash investigation with connected and automated vehicle data: Insights from a survey of law enforcement.* Journal of Transportation Safety & Security, (2025), *0*(0), 1-30. **https://doi.org/10.1080/19439962.2025.2462803**
20. OpenAI, Josh Achiam, Steven Adler, Sandhini Agarwal, Lama Ahmad, Ilge Akkaya, Florencia Leoni Aleman, Diogo Almeida, Janko Altenschmidt, Sam Altman, and Shyamal Anadkat, *GPT-4 Technical Report.* arXiv preprint, (2024). **https://doi.org/10.48550/arXiv.2303.08774**
21. Chiriatti, Massimo, Marianna Ganapini, Enrico Panai, Mario Ubiali, and Giuseppe Riva, *The case for human–AI interaction as system 0 thinking.* Nature Human Behaviour, (2024), *8*(10), 1829-1830. **https://doi.org/10.1038/s41562-024-01995-5**
22. Brown, Tom B., Benjamin Mann, Nick Ryder, Melanie Subbiah, Jared Kaplan, Prafulla Dhariwal, Arvind Neelakantan, Pranav Shyam, Girish Sastry, Amanda Askell, Sandhini Agarwal, Ariel Herbert-Voss, Gretchen Krueger, Tom Henighan, Rewon Child, Aditya Ramesh, Daniel M. Ziegler, Jeffrey Wu, Clemens Winter, Christopher Hesse, Mark Chen, Eric Sigler, Mateusz Litwin, Scott Gray, Benjamin Chess, Jack Clark, Christopher Berner, Sam McCandlish, Alec Radford, Ilya Sutskever, and Dario Amodei, *Language Models are Few-Shot Learners.* (2020). **https://doi.org/10.48550/arXiv.2005.14165**
23. Gemini Team, Google, *Gemini: A Family of Highly Capable Multimodal Models.* arXiv preprint, (2025). **https://doi.org/10.48550/arXiv.2312.11805**
24. Nie, Tong, Jian Sun, and Wei Ma, *Exploring the roles of large language models in reshaping transportation systems: A survey, framework, and roadmap.* Artificial Intelligence for Transportation, (2025), *1*, 100003. **https://doi.org/10.1016/j.ait.2025.100003**

25. Chen, Yanda, Joe Benton, Ansh Radhakrishnan, Jonathan Uesato, Carson Denison, John Schulman, Arushi Somani, Peter Hase, Misha Wagner, Fabien Roger, Vlad Mikulik, Sam Bowman, Jan Leike, Jared Kaplan, and Ethan Perez, *Reasoning Models Don't Always Say What They Think.* arXiv preprint, (2025). **https://doi.org/10.48550/arXiv.2505.05410**

26. DeepSeek-AI, Daya Guo, Dejian Yang, Haowei Zhang, Junxiao Song, Ruoyu Zhang, Runxin Xu, Qihao Zhu, Shirong Ma, Peiyi Wang, and Xiao Bi, *DeepSeek-R1: Incentivizing Reasoning Capability in LLMs via Reinforcement Learning.* arXiv preprint, (2025). **https://doi.org/10.48550/arXiv.2501.12948**

27. Wei, Jason, Xuezhi Wang, Dale Schuurmans, Maarten Bosma, Fei Xia, Ed Chi, Quoc V Le, and Denny Zhou, *Chain-of-thought prompting elicits reasoning in large language models.* NIPS'22: Proceedings of the 36th International Conference on Neural Information Processing System, (2022), *35*, 24824-24837. **https://doi.org/10.48550/arXiv.2201.11903**

28. Wu, Kebin, Wenbin Li, and Xiaofei Xiao, *AccidentGPT: Large Multi-Modal Foundation Model for Traffic Accident Analysis.* arXiv preprint, (2024). **https://doi.org/10.48550/arXiv.2401.03040**

29. Zhong, Wuchang, Jinglin Huang, Maoqiang Wu, Weinan Luo, and Rong Yu, *Large language model based system with causal inference and Chain-of-Thoughts reasoning for traffic scene risk assessment.* Knowledge-Based Systems, (2025), *319*, 113630. **https://doi.org/10.1016/j.knosys.2025.113630**

30. Asgari, Elham, Saleh Khalil, Nina Montaña-Brown, Magda Dubois, Jasmine Balloch, Joshua Au Yeung, and Dominic Pimenta, *A Framework to Assess Clinical Safety and Hallucination Rates of LLMs for Medical Text Summarisation.* (2024). **https://doi.org/10.1101/2024.09.12.24313556**

31. Omar, Mahmud, Vera Sorin, Jeremy D Collins, David Reich, Robert Freeman, Nicholas Gavin, Alexander Charney, Lisa Stump, Nicola Luigi Bragazzi, and Girish N Nadkarni, *Multi-model assurance analysis showing large language models are highly vulnerable to adversarial hallucination attacks during clinical decision support.* Communications Medicine, (2025), *5*(1), 330.

32. Singh, Sarvagya, Rahul Saha, Gulshan Kumar, Anand Nayyar, and Tae-Kyung Kim, *Are you hallucinated? Insights into large language models.* ICT Express, (2025), S2405959525002115. **https://doi.org/10.1016/j.icte.2025.12.011**

33. Schluntz, Erik and Barry Zhang. 2024. *Building Effective Agents.* Research at Anthropic (Anthropic). **https://www.anthropic.com/engineering/building-effective-agents**.

34. Julia Wiesinger, Patrick Marlow, Vladimir Vuskovic. 2025. *Agents White Paper.* Google (Google). **https://www.kaggle.com/whitepaper-agents**.

35. Xi, Zhiheng, Wenxiang Chen, Xin Guo, Wei He, Yiwen Ding, Boyang Hong, Ming Zhang, Junzhe Wang, Senjie Jin, Enyu Zhou, Rui Zheng, Xiaoran Fan, Xiao Wang, Limao Xiong, Yuhao Zhou, Weiran Wang, Changhao Jiang, Yicheng Zou, Xiangyang Liu, Zhangyue Yin, Shihan Dou, Rongxiang Weng, Wenjuan Qin, Yongyan Zheng, Xipeng Qiu, Xuanjing Huang, Qi Zhang, and Tao Gui, *The rise and potential of large language model based agents: a survey.* Science China Information Sciences, (2025), *68*(2). **https://doi.org/10.1007/s11432-024-4222-0**

36. Lu, Yikang, Alberto Aleta, Chunpeng Du, Lei Shi, and Yamir Moreno, *LLMs and generative agent-based models for complex systems research.* Physics of Life Reviews, (2024), *51*, 283-293. **https://doi.org/10.1016/j.plrev.2024.10.013**

37. Vrdoljak, Josip, Zvonimir Boban, Ivan Males, Roko Skrabic, Marko Kumric, Anna Ottosen, Alexander Clemencau, Josko Bozic, and Sebastian Völker, *Evaluating large language and large reasoning models* *as decision support tools in emergency internal medicine.* Computers in Biology and Medicine, (2025), *192*, 110351. **https://doi.org/10.1016/j.compbiomed.2025.110351**

38. Wu, Chaoyi, Pengcheng Qiu, Jinxin Liu, Hongfei Gu, Na Li, Ya Zhang, Yanfeng Wang, and Weidi Xie, *Towards evaluating and building versatile large language models for medicine.* npj Digital Medicine, (2025), *8*(1), 58. **https://doi.org/10.1038/s41746-024-01390-4**
39. Ashqar, Huthaifa I., Ahmed Jaber, Taqwa I. Alhadidi, and Mohammed Elhenawy, *Advancing Object Detection in Transportation with Multimodal Large Language Models (MLLMs): A Comprehensive Review and Empirical Testing.* Computation, (2025), *13*(6), 133. **https://doi.org/10.3390/computation13060133**
40. Li, Xinyi, Sai Wang, Siqi Zeng, Yu Wu, and Yi Yang, *A survey on LLM-based multi-agent systems: workflow, infrastructure, and challenges.* Vicinagearth, (2024), *1*(1), 9. **https://doi.org/10.1007/s44336-024-00009-2**
41. Gao, Mingyan, Yanzi Li, Banruo Liu, Yifan Yu, Phillip Wang, Ching-Yu Lin, and Fan Lai, *Single-agent or Multi-agent Systems? Why Not Both?* (2025). **https://doi.org/10.48550/arXiv.2505.18286**
42. Kitajima, Sou, Shimono Keisuke, Tajima Jun, Antona-Makoshi Jacobo, and Nobuyuki and Uchida, *Multi-agent traffic simulations to estimate the impact of automated technologies on safety.* Traffic Injury Prevention, (2019), *20*(sup1), S58-S64. **https://doi.org/10.1080/15389588.2019.1625335**
43. Chen, Junzhou and Sidi Lu, *An Advanced Driving Agent with the Multimodal Large Language Model for Autonomous Vehicles*, in *2024 IEEE International Conference on Mobility, Operations, Services and Technologies (MOST)*. 2024, IEEE: Dallas, TX, USA. p. 1-11.
44. Haji, Fatemeh, Mazal Bethany, Maryam Tabar, Jason Chiang, Anthony Rios, and Peyman Najafirad, *Improving LLM Reasoning with Multi-Agent Tree-of-Thought Validator Agent.* arXiv preprint, (2024). **https://doi.org/10.48550/arXiv.2409.11527**
45. Zhang, Miao, Zhenlong Fang, Tianyi Wang, Shuai Lu, Xueqian Wang, and Tianyu Shi, *CCMA: A framework for cascading cooperative multi-agent in autonomous driving merging using Large Language Models.* Expert Systems with Applications, (2025), *282*, 127717. **https://doi.org/10.1016/j.eswa.2025.127717**
46. Guo, Xusen, Qiming Zhang, Junyue Jiang, Mingxing Peng, Meixin Zhu, and Hao Frank Yang, *Towards explainable traffic flow prediction with large language models.* Communications in Transportation Research, (2024), *4*, 100150. **https://doi.org/10.1016/j.commtr.2024.100150**
47. Guo, Xusen, Xinxin Yang, Mingxing Peng, Hongliang Lu, Meixin Zhu, and Hai Yang, *Automating Traffic Model Enhancement with Ai Research Agent.* (2024). **https://doi.org/10.2139/ssrn.4995746**
48. Farquhar, Sebastian, Jannik Kossen, Lorenz Kuhn, and Yarin Gal, *Detecting hallucinations in large language models using semantic entropy.* Nature, (2024), *630*(8017), 625-630. **https://doi.org/10.1038/s41586-024-07421-0**
49. Lu, Ming Y., Bowen Chen, Drew F. K. Williamson, Richard J. Chen, Melissa Zhao, Aaron K. Chow, Kenji Ikemura, Ahrong Kim, Dimitra Pouli, Ankush Patel, Amr Soliman, Chengkuan Chen, Tong Ding, Judy J. Wang, Georg Gerber, Ivy Liang, Long Phi Le, Anil V. Parwani, Luca L. Weishaupt, and Faisal Mahmood, *A multimodal generative AI copilot for human pathology.* Nature, (2024), *634*(8033), 466-473. **https://doi.org/10.1038/s41586-024-07618-3**

50. Ali, Zafar, Yi Huang, Asad Khan, Guilin Qi, Yuxin Zhang, Junlan Feng, Chao Deng, and Pavlos Kefalas, *Pythia-RAG: Retrieval-augmented generation over a unified multimodal knowledge graph for enhanced QA.* Knowledge-Based Systems, (2026), *335*, 115200. **https://doi.org/10.1016/j.knosys.2025.115200**
51. NHTSA. 2025. *NHTSA Field Crash Investigation 2023 Coding and Editing Manual.* National Highway Traffic Safety Administration (Washington, DC, United States). **https://crashstats.nhtsa.dot.gov/Api/Public/ViewPublication/813614**.
52. Sandmann, Sarah, Stefan Hegselmann, Michael Fujarski, Lucas Bickmann, Benjamin Wild, Roland Eils, and Julian Varghese, *Benchmark evaluation of DeepSeek large language models in clinical decision-making.* Nature Medicine, (2025). **https://doi.org/10.1038/s41591-025-03727-2**
53. Yin, Shukang, Chaoyou Fu, Sirui Zhao, Ke Li, Xing Sun, Tong Xu, and Enhong Chen, *A survey on multimodal large language models.* National Science Review, (2024), *11*(12). **https://doi.org/10.1093/nsr/nwae403**
54. Behimehr, Sara and Hamid R. Jamali, *Relations between Cognitive Biases and Some Concepts of Information Behavior.* Data and Information Management, (2020), *4*(2), 109-118. **https://doi.org/10.2478/dim-2020-0007**
55. Acciarini, Chiara, Federica Brunetta, and Paolo Boccardelli, *Cognitive biases and decision-making strategies in times of change: a systematic literature review.* Management Decision, (2020), *59*(3), 638-652. **https://doi.org/10.1108/md-07-2019-1006**
56. Farber, Shai, *AI as a decision support tool in forensic image analysis: A pilot study on integrating large language models into crime scene investigation workflows.* Journal of Forensic Sciences, (2025), *70*(3), 932-943. **https://doi.org/10.1111/1556-4029.70035**
57. Kojima, Takeshi, Shixiang Shane Gu, Machel Reid, Yutaka Matsuo, and Yusuke Iwasawa, *Large Language Models are Zero-Shot Reasoners.* arXiv preprint, (2023). **https://doi.org/10.48550/arXiv.2205.11916**

# APPENDIX A: CASE STUDY 32548

To illustrate our AI-Agent driven framework performance on pre-crash analysis, we present a detailed analysis of one case study to show the workflow of our framework.

***Overview of Case study: 32548***

Case ID = 32548 involves a three-vehicle collision that occurred on January 27, 2023, at 17:03pm. At the time of the crash, lighting conditions were insufficient (dark) and the weather was cloudy. The crash involved three passenger vehicles: a Hyundai (Vehicle 1), a Toyota (Vehicle 2), and a Honda (Vehicle 3). The crash location was near an intersection.

According to the crash investigation records, Vehicle 2 continuously decelerated while approaching the intersection, reducing its speed from approximately 37 km/h to about 2 km/h. At the same time, Vehicle 3, which was traveling behind Vehicle 2, accelerated from approximately 56 km/h to about 64 km/h and subsequently rear-ended Vehicle 2, forming the first collision event (EVENT 1). After being struck from the rear, Vehicle 2 experienced a pronounced change in vehicle orientation and rotated toward the left. Its right side then collided laterally with the oncoming Vehicle 1, resulting in the second collision event (EVENT 2). The overall crash configuration is shown in Extended Data Figure. 1.

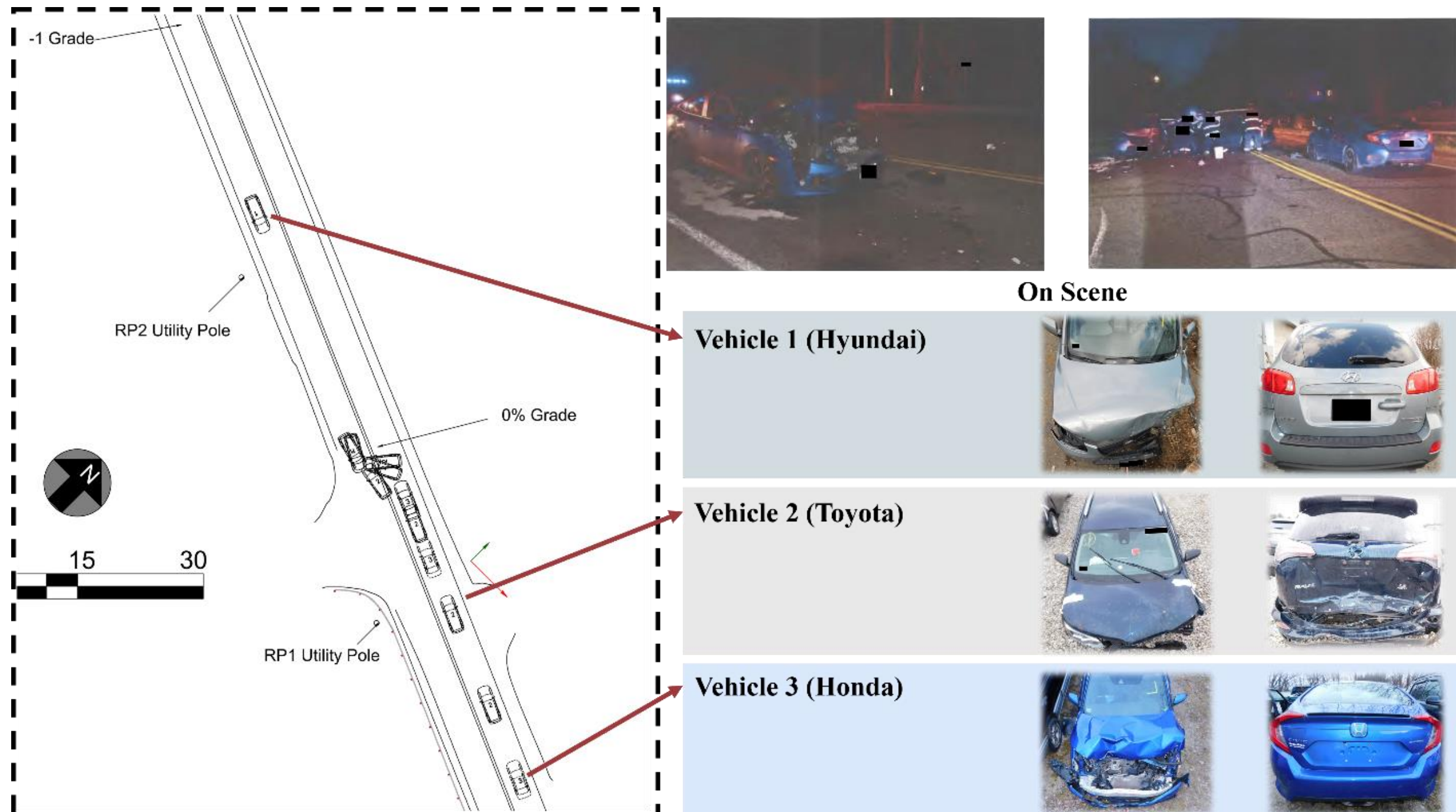

**Extended Data Figure. 1 Crash Overview of Case ID = 32548.**

Based on the directions of vehicle damage and the relative positions of the vehicles, Vehicle 3 is clearly identified as the striking vehicle and Vehicle 2 as the struck vehicle in the first collision event. Therefore, the subsequent analysis focuses on the EDR records of these two vehicles.

**For Vehicle 3**

Vehicle 3 recorded only one EDR event (EDREVENTNO1). This record shows that during the time interval from approximately −5 s to −1 s before the collision, vehicle speed increased continuously, the accelerator pedal remained engaged, and the brake pedal was not activated until less than 0.5 s before impact. This indicates that the vehicle was in an accelerating state for most of the period immediately preceding the collision.

**For Vehicle 2**

Vehicle 2 has six EDR records. Among them, EDREVENTNO6 is explicitly labeled by NHTSA as unrelated to this crash and is therefore excluded. Of the remaining five records, EDREVENTNO5 and EDREVENTNO2 are labeled as related to the first collision event, while EDREVENTNO1, EDREVENTNO3, and EDREVENTNO4 are labeled as related to the second collision event.

EDREVENTNO5 reflects a process of continuous deceleration while approaching an intersection. Considering the absence of a traffic signal, the steering behavior observed near the end of the record suggests that the driver was preparing to make a left turn into the intersection:

- From −4.9 s to 0 s, vehicle speed decreased smoothly from 24 km/h to about 3 km/h.
- The accelerator pedal was not engaged throughout this interval.
- The brake pedal was engaged during −4.9 to −0.9 s and again during −0.4 to 0 s.
- The steering wheel angle remained relatively stable from −4.9 s to −0.4 s and then increased to 75° before the collision.。

In contrast, EDREVENTNO2 shows pre-crash data that are identical to those in EDREVENTNO1, EDREVENTNO3, and EDREVENTNO4. These data indicate a sequence in which the vehicle first decelerated, then suddenly accelerated while the steering wheel angle rotated in both positive and negative directions, likely due to the influence of an impact, followed by another collision:

- From −4.6 s to −1.6 s, vehicle speed decreased from 37 km/h to 2 km/h, then increased to 26 km/h at −1.1 s and further to 31 km/h at 0 s.
- The accelerator pedal was not engaged from −4.6 s to −1.1 s, but remained fully engaged (100%) from −0.6 s to 0 s.
- The brake pedal was engaged from −4.6 s to −2.1 s and disengaged from −1.6 s to 0 s.
- The steering wheel angle was stable from −4.6 s to −1.6 s, increased to approximately 154.5° between −1.6 s and −1.1 s, and then rotated in the opposite direction.

As shown in **Extended Data Figure. 2**, the five EDR records from Vehicle 2 exhibit clear temporal overlap across multiple data channels. This overlap indicates that these records do not correspond to independent events but instead reflect a defined sequence of triggered recordings. Combined with the crash configuration, it can be inferred that the first collision was a rear-end impact. Prior to this initial collision, the struck vehicle could exhibit sustained deceleration. After being rear-ended, the vehicle would be expected to show transient acceleration and steering changes induced by the impact, and then become involved in the second lateral collision.

**a)**

| CASEID | Vehicle Number | EDREVENTNO | EDR Event | CDC Code | Event Number | Description | CDC Event |
|---|---|---|---|---|---|---|---|
| 32548 | 2 | 1 | Most Recent Event | 12FDEW02 | 2 | V1 Front vs V2 Front | Event # 2: V1 Front vs V2 Front - 12FDEW02 |
| 32548 | 2 | 2 | 1st Prior Event | 06BDEW02 | 1 | V2 Back vs V3 Front | Event # 1: V2 Back vs V3 Front - 06BDEW02 |
| 32548 | 2 | 3 | 2nd Prior Event | 12FDEW02 | 2 | V1 Front vs V2 Front | Event # 2: V1 Front vs V2 Front - 12FDEW02 |
| 32548 | 2 | 4 | 3rd Prior Event | 12FDEW02 | 2 | V1 Front vs V2 Front | Event # 2: V1 Front vs V2 Front - 12FDEW02 |
| 32548 | 2 | 5 | 4th Prior Event | 06BDEW02 | 1 | V2 Back vs V3 Front | Event # 1: V2 Back vs V3 Front - 06BDEW02 |
| 32548 | 2 | 6 | 5th Prior Event | | | | Event not related to this crash |
| 32548 | 3 | 1 | Event Record 1 | 12FDEW02 | 1 | V2 Back vs V3 Front | Event # 1: V2 Back vs V3 Front - 12FDEW02 |

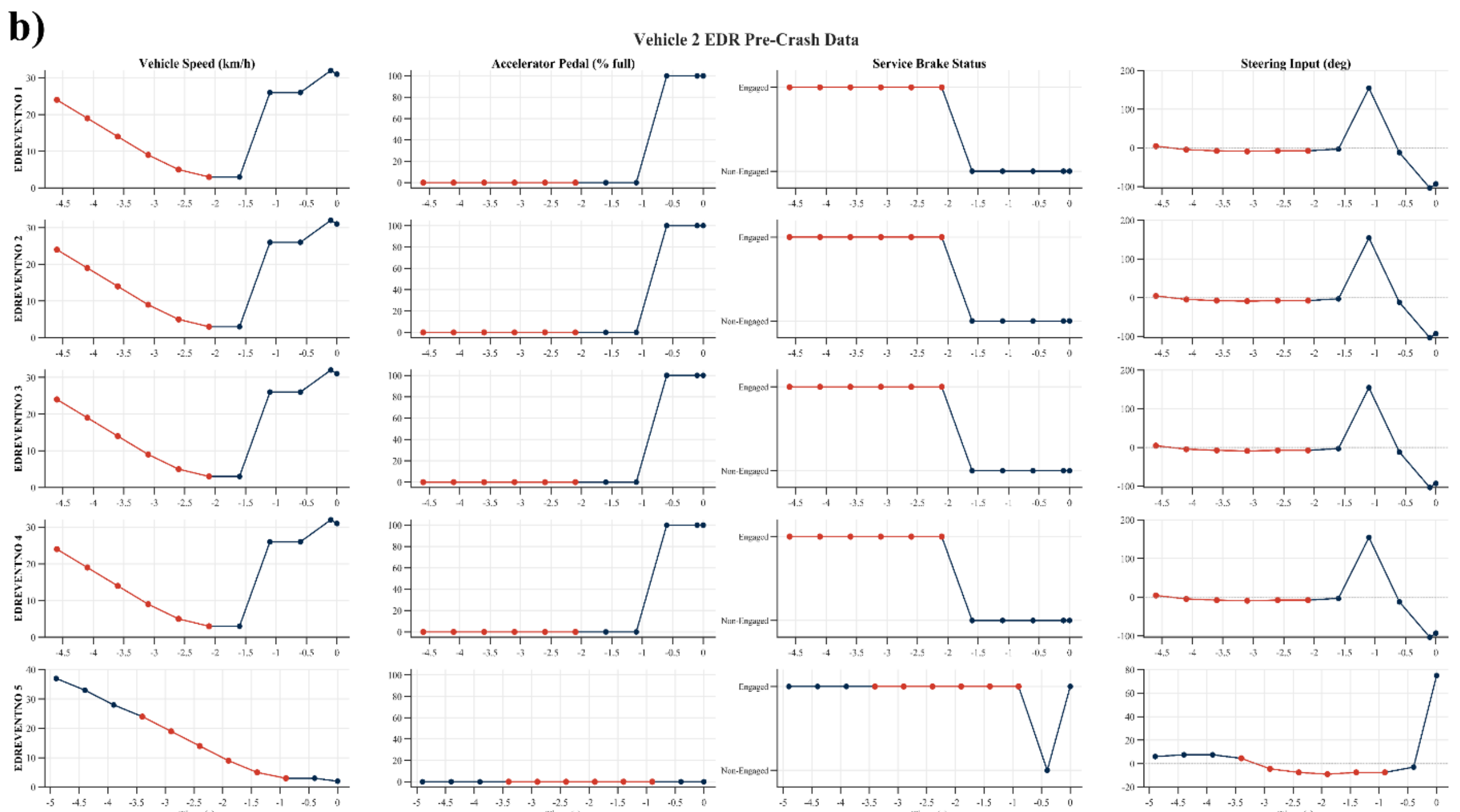

**Extended Data Figure. 2 EDR Source Data of Case ID = 32548.**

Among the available EDR records, only **EDREVENTNO5** fully captures a sustained deceleration pattern that is consistent with vehicle behavior prior to the initial rear-end collision. This record can therefore be identified as the one that most closely and accurately reflects the vehicle state immediately before the first collision event. Although **EDREVENTNO2** is labeled as related to EVENT 1, its data characteristics are more consistent with vehicle responses occurring after the second collision, suggesting that this record was more likely triggered by the second impact. This inconsistency may result from trigger uncertainty in the EDR recorder under multiple-impact conditions, or from disassociation of event numbers during the CISS data coding process. Regardless of the specific cause, the data in this case exhibit inherent and non-negligible internal inconsistencies.

***AI-Agent Framework in Case 32548***

The following shows the workflow of our AI-Agent Framework. In Case ID = 3254, the data will be transferred in to structured natural language format。Extended Data Figure. 3 shows the converted output for the Crash Scene Description, and Extended Data Figure. 4 shows the converted output for the EDR Data Analysis Report.

Crash Scene Description

# Total number of vehicles involved in this Crash: 3

# Vehicle Information:
## VEHNO=1
### Class of Vehicle: Small Passenger Car
### Damage Plane: Front
## VEHNO=2
### Class of Vehicle: Small Passenger Car
### Damage Plane: Front
## VEHNO=3
### Class of Vehicle: Unknown Vehicle Type
### Damage Plane: Front

# Crash Event Sequences Description:
EVENTNO1: Contact between Vehicle 2's back and Vehicle 3's front
EVENTNO2: Contact between Vehicle 1's front and Vehicle 2's front

# Environment Information:

## Environment for VEHNO=1:

### Crash event record 1 in this vehicle:
SPEEDLIMIT: 56 km/h
Trafficway Flow: Not physically divided (two-way traffic)
Travel Lanes: Two
Initial Travel Lane: One
Roadway alignment: Straight
Roadway Profile: Level (gradient <= 2%, vertical/horizontal)
Traffic Control: None present (no traffic control device in the area)
Pre-Event Movement (Prior to Recognition of Critical Event): Going straight
Critical Pre-Crash Event: Other Motor Vehicle Encroaching Into Lane: From opposite direction over left lane line (Source: External Vehicle Behavior - External factors, such as another vehicle encroaching or pedestrians, are the direct cause of the collision)
Crash Type: Miscellaneous → Backing, etc.: Other crash type

## Environment for VEHNO=2:

### Crash event record 1 in this vehicle:
SPEEDLIMIT: 56 km/h
Trafficway Flow: Not physically divided (two-way traffic)
Travel Lanes: Two
Initial Travel Lane: One
Roadway alignment: Straight
Roadway Profile: Level (gradient <= 2%, vertical/horizontal)
Traffic Control: None present (no traffic control device in the area)
Pre-Event Movement (Prior to Recognition of Critical Event): Decelerating in road
Critical Pre-Crash Event: Other Motor Vehicle in Lane: Traveling in same direction with higher speed (Source: External Vehicle Behavior - External factors, such as another vehicle encroaching or pedestrians, are the direct cause of the collision)
Crash Type: Same Trafficway, Same Direction → Rear-End: Decelerating (slowing), going left

## Environment for VEHNO=3:

### Crash event record 1 in this vehicle:
SPEEDLIMIT: 56 km/h
Trafficway Flow: Not physically divided (two-way traffic)
Travel Lanes: Two
Initial Travel Lane: One
Roadway alignment: Straight
Roadway Profile: Level (gradient <= 2%, vertical/horizontal)
Traffic Control: None present (no traffic control device in the area)
Pre-Event Movement (Prior to Recognition of Critical Event): Going straight
Critical Pre-Crash Event: Other Motor Vehicle in Lane: Traveling in same direction while decelerating (Source: External Vehicle Behavior - External factors, such as another vehicle encroaching or pedestrians, are the direct cause of the collision)
Crash Type: Same Trafficway, Same Direction → Rear-End: Decelerating (slowing)

# Notes:
1. Data Ownership: The encoded data is always based on the perspective of the subject vehicle, recording events that led to its collision.
2. Event Object: Event descriptions may involve the behavior of other vehicles from the subject vehicle's perspective, but the purpose of the data is to reflect the collision triggers encountered by the subject vehicle.
3. Multi-Vehicle Correlation: In the same accident, forms for different vehicles independently record their respective critical events, which may be interrelated (e.g., Vehicle A records Vehicle B's encroachment, while Vehicle B records its own loss of control).
3. This design ensures that each vehicle's pre-crash events are analyzed independently, while allowing the complete accident chain to be reconstructed through cross-referencing.

**Extended Data Figure. 3 Crash Scene Description of CaseID 32548.**

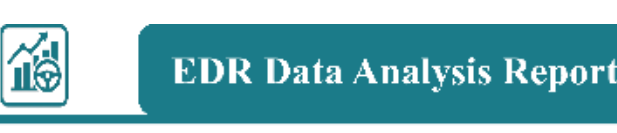

# EDR Data Analysis Report
## Basic description
This case (CASEID=32548) contains EDR data for 2 vehicles. In this EDR record, time zero (0 seconds) marks the triggering threshold of the recorded event for this vehicle, as determined by the EDR's deployment algorithm. The earliest data was recorded at -5.00 seconds (5.00 seconds pre-trigger), with the latest data at 0.00 seconds (0.00 seconds post-trigger). Time ranges may vary between vehicles due to differences in recording systems; consult 'vehicle_time_ranges' for per-vehicle specifics. Steering wheel angle values of 99996 (per SAE J1698) indicate invalid data and have been converted to null/N/A. [Note: EDR activation thresholds may include sudden deceleration, impact forces, or other vehicle dynamics parameters.]

### Summary Statistics
- CASEID: 32548
- num_vehicles: 2
- global_time_range: -5.00s to 0.00s

### Vehicle Time Ranges
- VEHNO2: -4.90s to 0.00s
- VEHNO3: -5.00s to 0.00s

## EDR Data for this Crash
### VEHNO2
#### EDREVENTNO1 (Prior Event 1)
##### Velocities
Speed of the vehicle in kilometers per hour. Sudden changes may indicate collision or evasive maneuvers.

| Time(sec) | Speed(kmph) | Notes |
| --------- | ----------- | ----- |
| -4.60 | 24.00 | |
| -4.10 | 19.00 | |
| -3.60 | 14.00 | |
| -3.10 | 9.00 | |
| -2.60 | 5.00 | |
| -2.10 | 3.00 | |
| -1.60 | 3.00 | |
| -1.10 | 26.00 | |
| -0.60 | 26.00 | |
| -0.10 | 32.00 | Peak speed |
| 0.00 | 31.00 | |

##### Steering Wheel Angles
Steering angle in degrees. Positive values indicate right turns, negative values indicate left turns.
Values near 0° indicate straight driving, while large changes indicate steering maneuvers.

| Time(sec) | Angle(deg) | Notes |
| --------- | ---------- | ----- |
| -4.60 | 4.50 | |
| -4.10 | -4.50 | |
| -3.60 | -7.50 | |
| -3.10 | -9.00 | |
| -2.60 | -7.50 | |
| -2.10 | -7.50 | |
| -1.60 | -3.00 | |
| -1.10 | 154.50 | Sharp left turn |
| -0.60 | -12.00 | Sharp right turn |
| -0.10 | -103.50 | Sharp right turn |
| 0.00 | -93.00 | Extreme steering angle |

##### Accelerator Pedal
Accelerator pedal position as percentage of full depression. 0% means pedal is not pressed, 100% means fully pressed.

| Time(sec) | Pedal(%) | Notes |
| --------- | -------- | ----- |
| -4.60 | 0.00 | |
| -4.10 | 0.00 | |
| -3.60 | 0.00 | |
| -3.10 | 0.00 | |
| -2.60 | 0.00 | |
| -2.10 | 0.00 | |
| -1.60 | 0.00 | |
| -1.10 | 0.00 | |
| -0.60 | 100.00 | Maximum acceleration input |
| -0.10 | 100.00 | |
| 0.00 | 100.00 | |

##### Brake Pedal Activation
Indicates whether brake pedal is activated. 'Yes' means brake is being applied, 'No' means brake is not applied.

| Time(sec) | Activated | Notes |
| --------- | --------- | ----- |
| -4.60 | Yes | Initial brake application |
| -4.10 | Yes | |
| -3.60 | Yes | |
| -3.10 | Yes | |
| -2.60 | Yes | |
| -2.10 | Yes | |
| -1.60 | No | Brake released |
| -1.10 | No | |
| -0.60 | No | |
| -0.10 | No | |
| 0.00 | No | |

..................................................................................

### VEHNO3
#### EDREVENTNO1 (Prior Event 1)
##### Velocities
Speed of the vehicle in kilometers per hour. Sudden changes may indicate collision or evasive maneuvers.

| Time(sec) | Speed(kmph) | Notes |
| --------- | ----------- | ----- |
| -5.00 | 56.00 | |
| -4.50 | 56.00 | |
| -4.00 | 58.00 | |
| -3.50 | 59.00 | |
| -3.00 | 60.00 | |
| -2.50 | 61.00 | |
| -2.00 | 63.00 | |
| -1.50 | 63.00 | |
| -1.00 | 64.00 | Peak speed |
| -0.50 | 64.00 | |
| 0.00 | 64.00 | |

##### Steering Wheel Angles
..................................................................................

## CDC and EDR Event Description (Most crucial from NHSTA investigation report)
This section connects EDR events to the physical collision events identified in the investigation.
For VEHNO=2, EDREVENTNO=1, corresponds to EVENTNO2: V1 Front vs V2 Front
For VEHNO=2, EDREVENTNO=2, corresponds to EVENTNO1: V2 Back vs V3 Front
For VEHNO=2, EDREVENTNO=3, corresponds to EVENTNO2: V1 Front vs V2 Front
For VEHNO=2, EDREVENTNO=4, corresponds to EVENTNO2: V1 Front vs V2 Front
For VEHNO=2, EDREVENTNO=5, corresponds to EVENTNO1: V2 Back vs V3 Front
For VEHNO=2, EDREVENTNO=6, corresponds to EVENTNO: Event not related to this crash
For VEHNO=3, EDREVENTNO=1, corresponds to EVENTNO1: V2 Back vs V3 Front

**Extended Data Figure. 4 EDR Data Analysis Report of CaseID 32548.** Due to the length, only the sample data format for Vehicle 2's EDR EVENT NO1 is fully provided here. The complete EDR Data Analysis Report of this case is 2,866 words.

During model interaction, all data are passed to the large language model through prompts. A prompt is the complete textual input the model receives at runtime. It contains both the task instructions and all background information relevant to the task.

In Phase I, the prompt—embedding the Crash Scene Description and the Scene Diagram—is provided to Agent 1. Based on this full input, the model produces an overall description of the crash process and generates an Accident Reconstruction Report (see Extended Data Figure 5 for details).

In Phase II, the Accident Reconstruction Report generated in Phase I and the EDR Data Analysis Report are embedded together into a new prompt and passed to Agent 2. Agent 2 then identifies the crash-involved party that is most relevant to the initial collision event and provides the corresponding rationale.

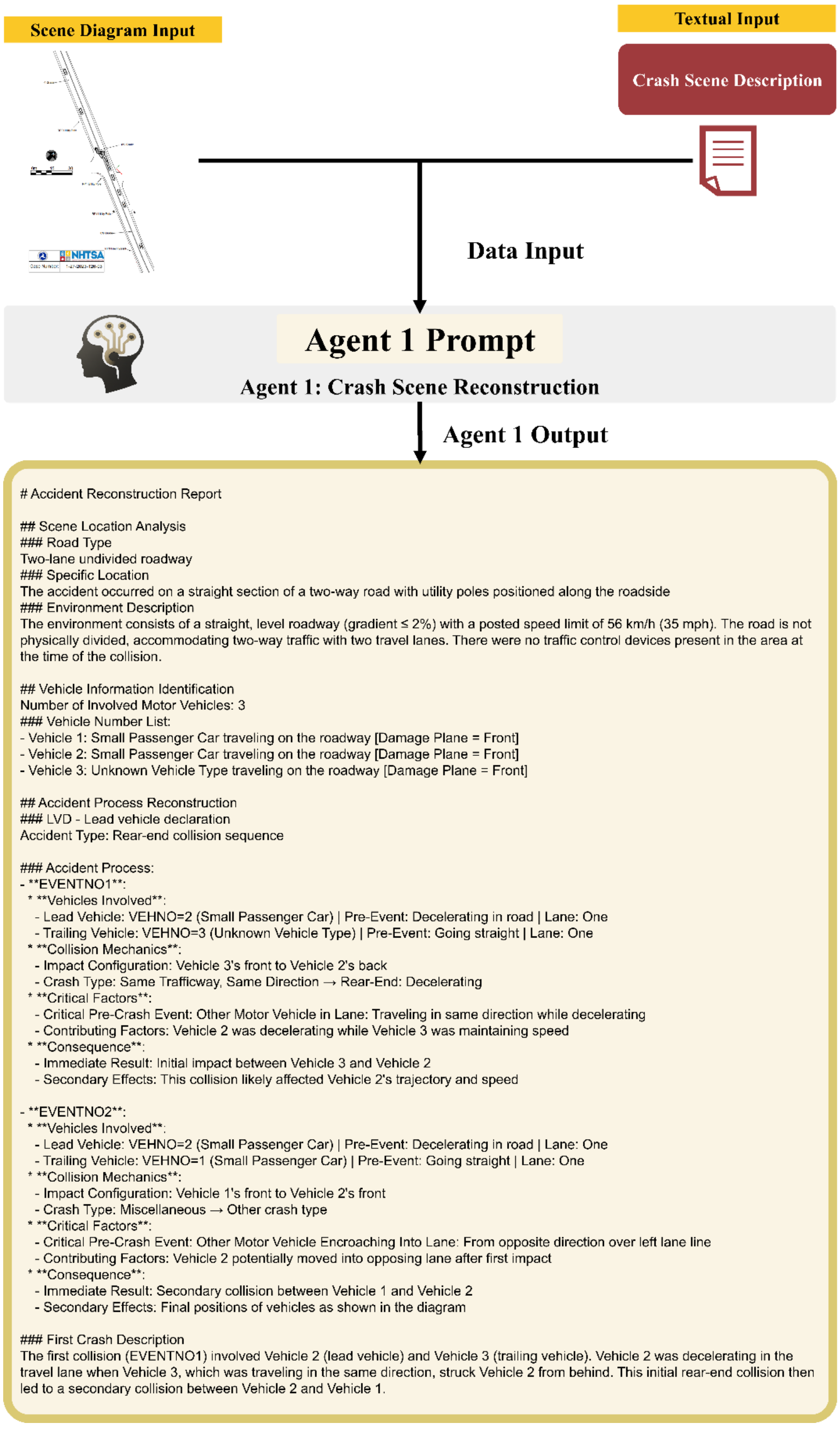

**Extended Data Figure. 5 Phase I AI-Agent Processing.** Illustration of the complete workflow showing how structured crash scene description data and scene diagrams are embedded into the prompt processed by Agent 1, and generate the output.

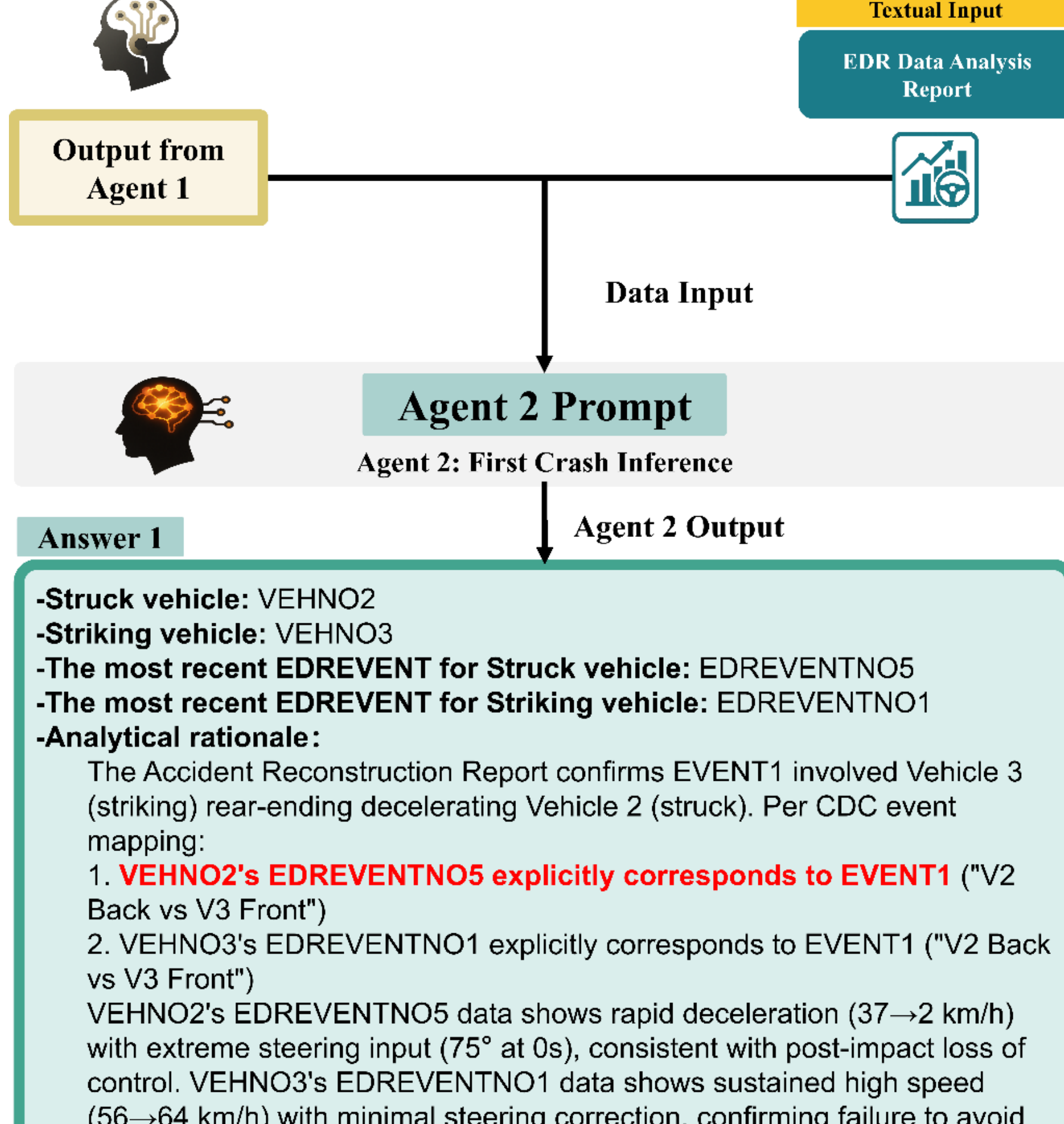

**Extended Data Figure. 6 Phase II AI-Agent Processing.** Illustration of how Agent 1's output (Accident Reconstruction Report) is combined with the EDR Data Analysis Report, integrated through the prompt, and transmitted to Agent 2 to generate the First Crash Inference conclusion.

As shown in Extended Data Figure. 6, under identical inputs, Agent 2 produced consistent final decisions across multiple runs, although the supporting reasoning was not identical. While maintaining the same conclusion, different outputs emphasized different evidence combinations and exclusion logic. Notably, Answer 2 explicitly states that the time-series characteristics of EDREVENTNO2 are inconsistent with the initial rear-end collision: the speed rebound and associated dynamics are more consistent with the vehicle participating in a second lateral collision after the first impact. Therefore, EDREVENTNO2 is excluded as the record corresponding to the initial collision event.

This case illustrates that, even when the final conclusion remains the same, a large language model may follow different reasoning paths to detect and articulate internal inconsistencies in the data, thereby avoiding or reducing the influence of misleading information on the final judgment.